\documentclass[pdflatex,iicol,sn-nature,Numbered]{sn-jnl}
\usepackage[T1]{fontenc}

\usepackage{graphicx}
\usepackage{booktabs}
\usepackage{longtable}
\usepackage{array}
\usepackage{tabularx}
\usepackage{calc}
\usepackage{amsmath,amssymb}
\usepackage{xurl}
\usepackage{placeins}
\graphicspath{{figures/}}

\begin{document}

\title[A HamNoSys-Guided Handshape Dataset and Benchmark]
{A HamNoSys-Guided Dataset and Baselines for Fine-Grained Isolated Handshape Recognition in Sign Language}

\author*[1,2]{\fnm{Ushnish} \sur{Sarkar}}\email{u.sarkar@vecc.gov.in}
\author[3]{\fnm{Suvajit} \sur{Patra}}
\equalcont{These authors contributed equally to this work.}
\author[1,2]{\fnm{Bhaswar} \sur{Chattopadhyay}}
\equalcont{These authors contributed equally to this work.}
\author[1]{\fnm{Pranab} \sur{Singha Roy}}
\equalcont{These authors contributed equally to this work.}
\author[1,2]{\fnm{Tapas} \sur{Samanta}}
\equalcont{These authors contributed equally to this work.}

\affil*[1]{\orgdiv{Computer and Informatics Group},
  \orgname{Variable Energy Cyclotron Centre},
  \orgaddress{\city{Kolkata}, \postcode{700064}, \country{India}}}
\affil[2]{\orgname{Homi Bhabha National Institute},
  \orgaddress{\city{Mumbai}, \postcode{400094}, \country{India}}}
\affil[3]{\orgname{Ramakrishna Mission Vivekananda Educational and Research Institute},
  \orgaddress{\city{Belur}, \postcode{711202}, \country{India}}}

\abstract{\textbf{Purpose:} Fine-grained handshape recognition supports computational sign-language transcription, recognition, and translation, but broad, phonetically defined visual inventories with signer-aware evaluation remain limited. This work introduces a handshape dataset  grounded in the language-independent Hamburg Notation System (HamNoSys) and baseline models for handshape recognition evaluated on the same.

\textbf{Methods:} A balanced dataset of 144,000 RGB images was collected from 15 participants for 160 handshape classes defined by the official HamNoSys 4 Handshapes Chart. ResNet-18 and ViT-B/16 as appearance-based models were evaluated on this dataset, while a graph convolutional network and XGBoost were evaluated on the hand landmarks of the images from this dataset. Both a class-stratified subject-dependent split and a 15-fold leave-one-subject-out (LOSO) protocol were used. The same model families were additionally assessed on LSWH100 and ASL Fingerspelling Dataset A for external context.

\textbf{Results:} The subject-dependent baselines established reproducible reference performance across all four model families, whereas LOSO evaluation showed some reduction when recognition was required to generalise to unseen participants. Additional analysis has been shown to highlight the presence of visually close handshapes that may possibly confuse the  standard models .

\textbf{Conclusion:} The documented acquisition, curation, and complementary evaluation protocols provide a reproducible resource for fine-grained isolated-handshape research and for developing more accessible sign-language technologies. }


\maketitle

\section{Introduction}\label{sec:introduction}

More than 1.5 billion people worldwide experience some degree of
hearing loss \cite{who2024}, while a national survey in the United States
estimated that approximately 2.8\% of adults use a sign language
\cite{mitchell2023}. Accessible computational tools for sign-language
transcription and recognition are therefore relevant to a substantial
and diverse population.
Sign languages are natural languages with their own phonological,
morphological, and syntactic structures \cite{sandler2006}. At the
sublexical level, signs can be analysed through contrastive manual
parameters such as handshape, orientation, location, and movement, together
with linguistically relevant non-manual components
\cite{stokoe1960,sandler2006}. Among the manual parameters, handshape
provides an important source of lexical contrast and is consequently
relevant to computational sign-language recognition and translation
\cite{rastgoo2021}. Linguistically grounded representation, resource
construction, and evaluation have therefore remained central concerns in
sign-language technology \cite{bragg2019,decoster2024}.

The visual--gestural modality presents a fundamental representational
difficulty. Video preserves the multidimensional and temporally organised
signal, but it does not by itself provide an explicit, discrete, and
searchable description of the articulatory form. Glosses can be used to
identify lexical sign types or meanings, but differences in handshape,
orientation, location, movement, and signer-specific realisation are not
encoded by them. When physical sign form, phonetic variation, phonological
contrast, or cross-linguistic similarity is to be investigated, a
form-based transcription system is therefore required. Through such a
system, observable sublexical properties can be represented consistently,
sign variants can be compared, corpus annotations can be searched, and
findings can be evaluated across datasets and sign languages
\cite{garcia2013,tkachman2016}. However, no universally accepted
sign-language equivalent of the International Phonetic Alphabet has yet
been established \cite{tkachman2016}.

Several transcription systems have been developed at different levels of
descriptive abstraction. Stokoe notation introduced a parameter-based
phonological representation of handshape, location, and movement,
principally for American Sign Language \cite{stokoe1960}. The
Liddell--Johnson Movement--Hold model subsequently represented signs
through temporally ordered movement and hold segments
\cite{liddell1989}. Prosodic Model Handshape Coding provides a
theoretically motivated phonological representation of selected fingers
and joint configuration \cite{eccarius2008}, whereas Sign Language
Phonetic Annotation provides a more anatomically detailed description of
individual fingers, joints, thumb configuration, and articulator contact
\cite{johnson2011,tkachman2016,hall2017}. SignWriting serves primarily as
a graphical or orthographic representation and has also been employed as
an intermediate representation in computational translation
\cite{jiang2023}. These systems differ in purpose, descriptive coverage,
and granularity and are therefore not interchangeable
\cite{hochgesang2014, dhanjal2019}.
For the present handshape-centred task, HamNoSys offers greater
articulatory detail and broader cross-linguistic applicability than the
original Stokoe system, while avoiding the extensive joint-by-joint
description required by Sign Language Phonetic Annotation
\cite{hochgesang2014,hall2017}. Unlike a notation restricted to a
particular phonological model, it can also be used as a mostly phonetic
description of observable manual form \cite{prillwitz1989,hanke2004}. Most importantly for dataset construction, the official, non-exhaustive HamNoSys 4 Handshapes Chart
provides publicly illustrated reference forms from which a bounded and
visually reproducible class inventory can be defined
\cite{hanke2010chart}. Reported inconsistencies in the use of HamNoSys
annotations further support the use of labels tied directly to fixed
chart illustrations rather than unconstrained transcription
\cite{ferlin2024}.
Existing visual resources support several related tasks, including
fingerspelling recognition, corpus-derived handshape classification, and
synthetic handshape recognition. However, their differing objectives
leave a need for a real-image benchmark that combines a broad,
transcription-defined handshape inventory with systematic evaluation on
unseen participants. The supporting comparison with existing resources
is provided in Section~\ref{sec:related-work}.
This need was addressed through the construction of a balanced,
HamNoSys-grounded image dataset and its evaluation under complementary
subject-dependent and leave-one-subject-out protocols. The class
definition, acquisition procedure, and dataset composition are presented
in Section~\ref{sec:dataset}, while the baseline models and matched-model
experiments on external datasets are presented in
Section~\ref{sec:experiments}.
\section{Related Work}
\label{sec:related-work}

Fine-grained handshape inventories distinguish selected fingers, joint
configuration, thumb behaviour, inter-finger relations, and contact
\cite{eccarius2008,brentari2010,brentari2017}. Segmental descriptions
additionally distinguish static articulation from its temporal
organisation within a sign \cite{johnson2011}. Because neighbouring
handshapes may differ in only one of these properties, fine-grained
recognition requires a broader inventory than that provided by
language-specific fingerspelling alphabets.

Existing visual resources represent several related but distinct tasks.
Continuous-sign corpora provide weak or auxiliary handshape labels
\cite{koller2016,zhang2023handshape}; LSWH100 contains synthetic images
organised using SignWriting-derived categories \cite{loboneto2024};
and fingerspelling datasets cover restricted language-specific
alphabets or digits
\cite{signLanguageMNIST2017,pugeault2011,hosoe2017,jain2023}.
Other resources represent complete lexical signs
\cite{ronchetti2016} or derive hand-pose categories through
feature-based clustering \cite{kajiyama2022}. Their differences in
visual data, class inventory, and linguistic scope are summarised in
Table~\ref{tab:handshape-datasets}.
\begin{table*}[!t]
\centering
\footnotesize
\caption{Representative visual handshape and sign-language resources.}
\label{tab:handshape-datasets}
\setlength{\tabcolsep}{4pt}
\renewcommand{\arraystretch}{1.10}
\begin{tabularx}{\textwidth}{@{}
>{\raggedright\arraybackslash}p{3.05cm}
>{\raggedright\arraybackslash}p{3.35cm}
>{\centering\arraybackslash}p{1.35cm}
>{\raggedright\arraybackslash}X@{}}
\toprule
Resource & Visual data & Classes & Label basis or scope \\
\midrule
Deep Hand \cite{koller2016} &
$>1$ million weakly labelled real frames &
60 &
Corpus- and lexicon-derived handshapes \\
\addlinespace
PHOENIX14T-HS \cite{zhang2023handshape} &
Continuous-sign videos &
60 &
Auxiliary handshape labels associated with DGS glosses \\
\addlinespace
LSWH100 \cite{loboneto2024} &
144,000 synthetic images &
100 &
SignWriting-derived Libras handshapes \\
\addlinespace
ASL Fingerspelling Dataset A \cite{pugeault2011} &
65,774 real RGB images from five users &
24 &
Static ASL alphabet \\
\addlinespace
Other fingerspelling resources
\cite{signLanguageMNIST2017,hosoe2017,jain2023} &
Static real or augmented images &
24--41 &
ASL, JSL, or Danish letters and digits \\
\addlinespace
LSA64 \cite{ronchetti2016} &
3,200 videos &
64 &
Complete Argentinian Sign Language lexical signs \\
\addlinespace
Kajiyama et al.\ \cite{kajiyama2022} &
Hand images from sign-language data &
Data-derived &
Finger-shape and palm-orientation clusters \\
\addlinespace
Proposed dataset &
144,000 real RGB images from 15 participants &
160 &
Chart-defined HamNoSys handshapes \\
\bottomrule
\end{tabularx}
\end{table*}
Within the reviewed literature, no resource jointly provides an
approximately balanced collection of real isolated-hand images, a broad
class inventory grounded in a language-independent phonetic notation,
and evaluation under both participant-overlapping and
participant-independent protocols. This combination defines the
specific resource gap addressed in the present study.
HamNoSys has otherwise been applied primarily at the sign level in
corpora, multilingual lexicons, dictionary transcription,
cross-linguistic comparison, and automatic motion generation
\cite{hanke2004,hanke2020dgs,konrad2007glex,matthes2012,lacheta2016,
villamonedero2023,varanasi2025}. As summarised in
Table~\ref{tab:hamnosys-resources}, these applications encode lexical
citation forms, corpus units, or motion sequences rather than balanced
image collections organised by isolated handshape. The HamNoSys 4
Handshapes Chart is itself an illustrated reference inventory rather
than an image dataset; its use for defining the proposed classes is
described in Section~\ref{sec:dataset}.

\begin{table*}[!t]
\centering
\footnotesize
\caption{Documented applications of HamNoSys in sign-language resources.}
\label{tab:hamnosys-resources}
\setlength{\tabcolsep}{4pt}
\renewcommand{\arraystretch}{1.10}
\begin{tabularx}{\textwidth}{@{}
>{\raggedright\arraybackslash}p{3.35cm}
>{\raggedright\arraybackslash}p{2.65cm}
>{\raggedright\arraybackslash}p{3.15cm}
>{\raggedright\arraybackslash}X@{}}
\toprule
Resource & Sign language(s) & Scale or primary unit & Function of HamNoSys \\
\midrule
DGS Corpus and GLex
\cite{hanke2004,hanke2020dgs,konrad2007glex} &
German Sign Language &
Corpus utterances and technical lexical signs &
Corpus-linked annotation and citation-form description \\
\addlinespace
DICTA-SIGN resources \cite{matthes2012,varanasi2025} &
BSL, DGS, GSL, and LSF &
Approximately 1,000 signs per language &
Citation-form representation and cross-linguistic comparison \\
\addlinespace
Corpus-based Dictionary of PJM \cite{lacheta2016} &
Polish Sign Language &
3,476 lexical signs &
Citation-form transcription \\
\addlinespace
Motion-generation dataset \cite{villamonedero2023} &
Spanish Sign Language &
754 signs and 6,786 videos &
Intermediate representation for automatic motion generation \\
\bottomrule
\end{tabularx}
\end{table*}

More broadly, gesture-recognition research has addressed
human--computer interaction, virtual environments, rehabilitation, and
sign-language processing \cite{mitra2007}. General-purpose gesture
resources, however, are not necessarily organised as linguistically
defined handshape inventories.

Complementary representation families have been adopted for handshape
and gesture recognition. RGB images have been processed using
convolutional and transformer-based architectures
\cite{he2016,dosovitskiy2021}, while hand landmarks have been represented
as anatomical graphs or fixed-length feature vectors
\cite{zhang2020mediapipe,kipf2017,sarkar24,chen2016}. Representative
baselines from these families are evaluated in the present study rather
than an exhaustive set of architectures.

Evaluation design is particularly important for multi-participant data.
Participant-overlapping splits measure recognition when the same
participants may occur across partitions, whereas
participant-independent protocols assess generalisation to unseen
individuals \cite{aly2020}. Both settings are therefore reported, with
leave-one-subject-out evaluation used for the systematic
participant-independent assessment described in
Section~\ref{sec:experiments}.

\section{Dataset Construction}
\label{sec:dataset}

This section defines the handshape class inventory and subsequently
describes the acquisition protocol, collection software, participants,
dataset organisation, and evaluation splits.

\subsection{HamNoSys Handshape Inventory and Class Definition}
\label{subsec:hamnosys-handshape-inventory-and-class-definition}

HamNoSys Version 2.0 encodes the manual components of a sign through
handshape, orientation, location, and movement \cite{prillwitz1989}.
Version 4 additionally supports optional non-manual specifications
\cite{hanke2004}. Handshapes are constructed compositionally from a
basic form and modifiers describing finger selection, bending, thumb
position, individual-finger configuration, and intermediate forms.
Transitions between handshapes are represented as actions rather than
separate dynamic handshape primitives \cite{hanke2004}.

\begin{figure*}[!t]
  \centering
  \includegraphics[width=\linewidth]
    {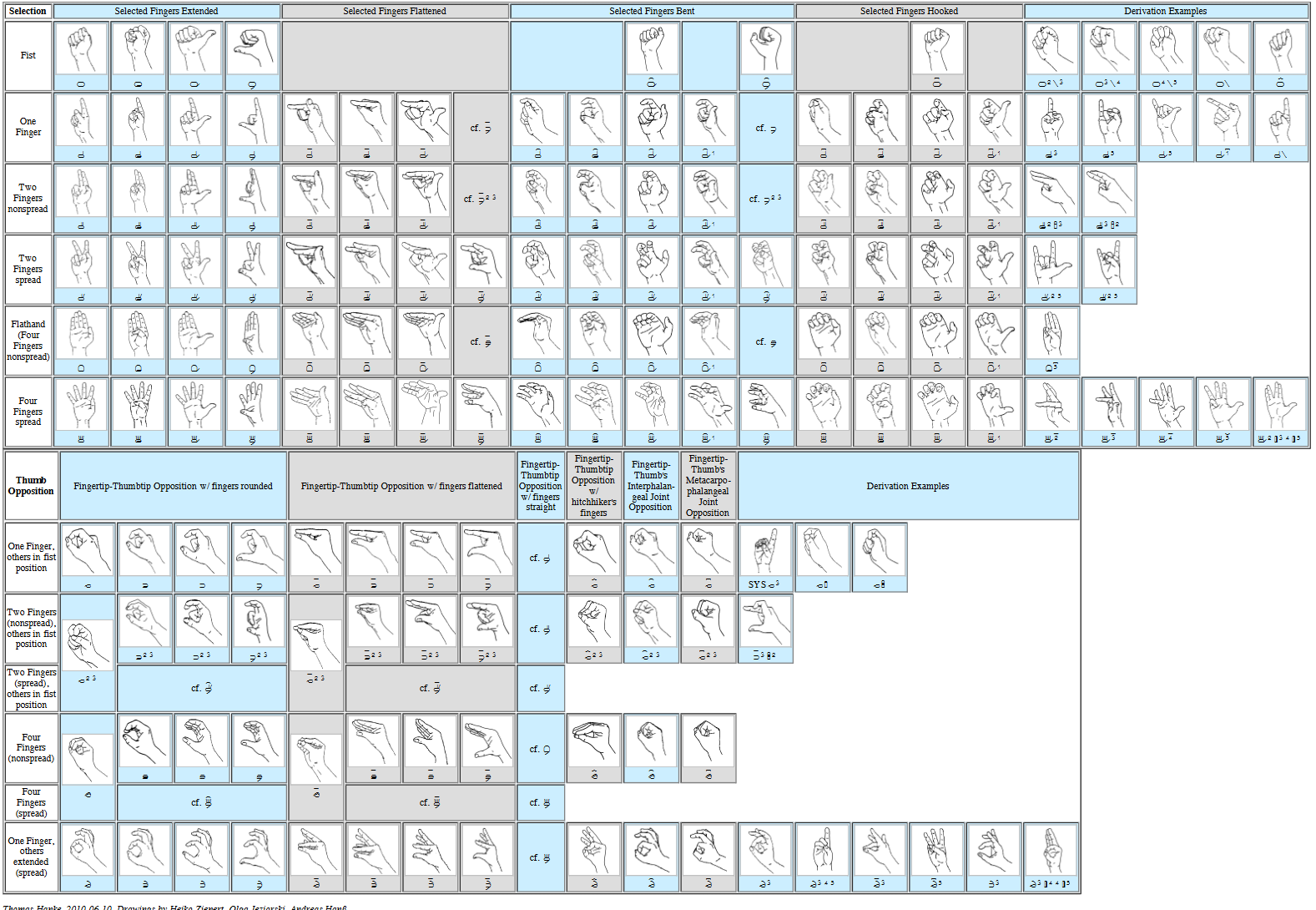}
  \caption{The official, non-exhaustive HamNoSys 4 Handshapes Chart
  \cite{hanke2010chart}. Each distinct drawn hand model is treated as one
  dataset class.}
  \label{fig:hamnosys-chart}
\end{figure*}

Because HamNoSys does not define a finite handshape inventory, the
official, non-exhaustive HamNoSys 4 Handshapes Chart was used to establish
a bounded and reproducible class set \cite{hanke2010chart}. Each distinct
illustrated hand model in Fig.~\ref{fig:hamnosys-chart} was treated as one
class; blank cells and cells containing only symbols or cross-references
were excluded. This procedure produced 160 classes.

The illustrations occupy six Selection rows and four Thumb-opposition
rows. The two Thumb-opposition rows containing no illustrations---\emph{Two
Fingers (spread), others in fist position} and \emph{Four Fingers
(spread)}---were excluded. Table~\ref{tab:selection-inventory} reports
the class counts for the ten populated rows, while
Table~\ref{tab:chart-group-counts} groups the same 160 classes by chart
column: 106 Selection classes and 54 Thumb-opposition classes. All dataset
codes are author-defined and are not official HamNoSys symbols.

\begin{table*}[!t]
\centering
\small
\caption{Populated chart selections used as dataset categories.}
\label{tab:selection-inventory}
\renewcommand{\arraystretch}{1.08}
\begin{tabularx}{\linewidth}{@{}lXcr@{}}
\toprule
Chart section & Category & Code & Classes \\
\midrule
Selection
  & Fist
  & F & 12 \\
Selection
  & One Finger
  & OF & 20 \\
Selection
  & Two Fingers (nonspread)
  & TFN & 17 \\
Selection
  & Two Fingers (spread)
  & TFS & 19 \\
Selection
  & Flathand (Four Fingers nonspread)
  & FFN & 16 \\
Selection
  & Four Fingers (spread)
  & FFS & 22 \\
\midrule
Thumb opposition
  & One Finger, others in fist position
  & OFO & 14 \\
Thumb opposition
  & Two Fingers (nonspread), others in fist position
  & TFO & 12 \\
Thumb opposition
  & Four Fingers (nonspread)
  & FFO & 11 \\
Thumb opposition
  & One Finger, others extended (spread)
  & OFOE & 17 \\
\midrule
\multicolumn{3}{r}{\textbf{Total}} & \textbf{160} \\
\bottomrule
\end{tabularx}
\end{table*}

\begin{table*}[!t]
\centering
\small
\caption{Number of drawn models under each chart column group.}
\label{tab:chart-group-counts}
\renewcommand{\arraystretch}{1.08}
\begin{tabularx}{\linewidth}{@{}lXcr@{}}
\toprule
Chart section & Column group & Code & Classes \\
\midrule
Selection & Selected Fingers Extended & SFE& 24 \\
Selection & Selected Fingers Flattened &SFF& 17 \\
Selection & Selected Fingers Bent & SFB& 24 \\
Selection & Selected Fingers Hooked & SFH &21 \\
Selection & Derivation Examples & DE & 20 \\
\midrule
Thumb opposition & Fingertip-Thumbtip Opposition w/fingers rounded & FTR & 16 \\
Thumb opposition & Fingertip-Thumbtip Opposition w/fingers flattened & FTF  &16 \\
Thumb opposition & Fingertip-Thumbtip Opposition w/Hitchhiker's fingers & FTH &4 \\
Thumb opposition & Fingertip Thumb's Interphalangeal-joint opposition & FTI &4 \\
Thumb opposition & Fingertip Thumb's Metacarpophalangeal-joint opposition & FTM &4 \\
Thumb opposition & Other Derivation Examples & DE & 10 \\
\midrule
\multicolumn{3}{r}{\textbf{Total}} & \textbf{160} \\
\bottomrule
\end{tabularx}
\end{table*}

The resulting inventory is restricted to the forms illustrated in the
non-exhaustive chart and therefore does not cover every handshape
expressible in HamNoSys. Extension beyond these 160 classes would require
expert definition and validation.

\subsection{Participants and Ethics}
\label{subsec:participants-and-ethics}

Fifteen university students aged 23--25 years participated in the data
collection. Right-hand dominance was reported by fourteen participants
and left-hand dominance by one. For each class, participants examined and
reproduced the displayed reference handshape. Finger selection, bending,
thumb position, and contact were verified against the reference before
recording by an operator with sign-language experience. Participants
received an honorarium for their time. The applicable ethical oversight
and consent procedures are reported in the Statements and Declarations.

\subsection{Dataset recording}
\label{subsec:acquisition-system}

\begin{figure*}[!t]
\centering
\includegraphics[width=0.98\linewidth]
  {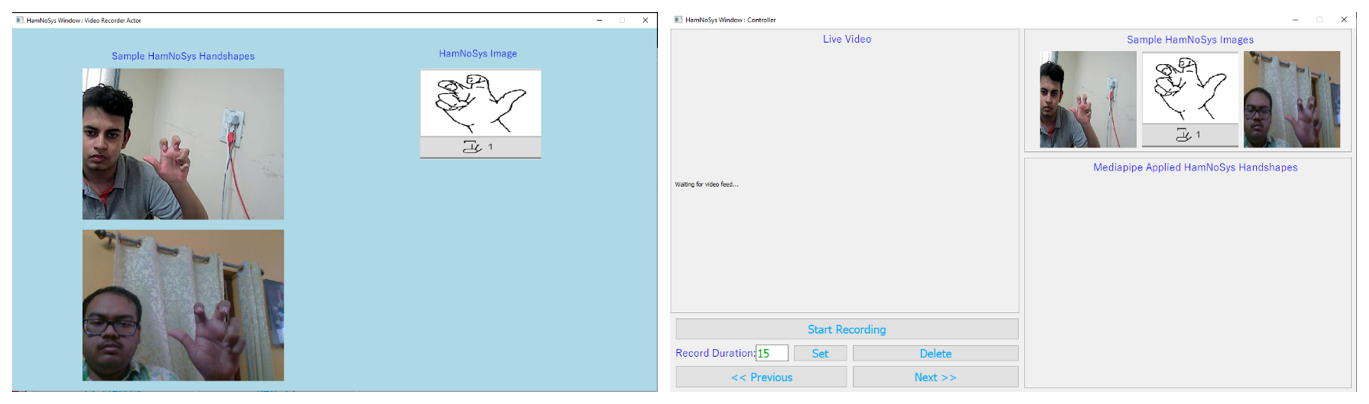}
\caption{Custom data-acquisition software: (a) actor interface displaying
the target handshape and live participant view, and (b)
operator/controller interface used for verification and recording
control.}
\label{fig:capture-software}
\end{figure*}

Two interfaces were provided by the acquisition application
(Fig.~\ref{fig:capture-software}). The target HamNoSys handshape and live
camera view were displayed on the actor panel, while the reference,
incoming stream, and recording controls were displayed on the operator
panel. The recording duration was set to 10 seconds.

\begin{figure*}[!t]
\centering
\includegraphics[width=0.92\linewidth]
  {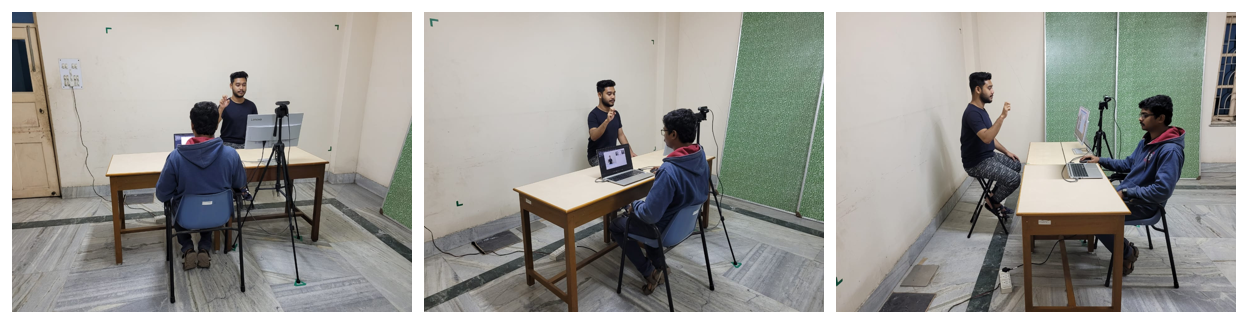}
\caption{Controlled classroom recording environment showing the
participant, operator, acquisition computer, and tripod-mounted RGB
camera from three viewpoints.}
\label{fig:recording-setup}
\end{figure*}

Recordings were made indoors against a constant dull-white background
using a tripod-mounted Logitech Brio RGB camera at 30 frames/s and
$640\times480$ pixels. The acquisition program was run on an Intel Core
i7 Windows workstation, while the session was managed from a separate
workstation (Fig.~\ref{fig:recording-setup}).

No instrumented gloves or body-mounted sensors were used. The recording
area was kept reasonably uncluttered to limit irrelevant background
variation and to support later hand localisation. Participants were free
to adjust their upper-body posture and hand position within the camera
view. The only required actions were to form the target handshape and
perform the prescribed rotation of the hand.

\begin{figure*}[t]
\centering
\includegraphics[width=0.98\linewidth]
  {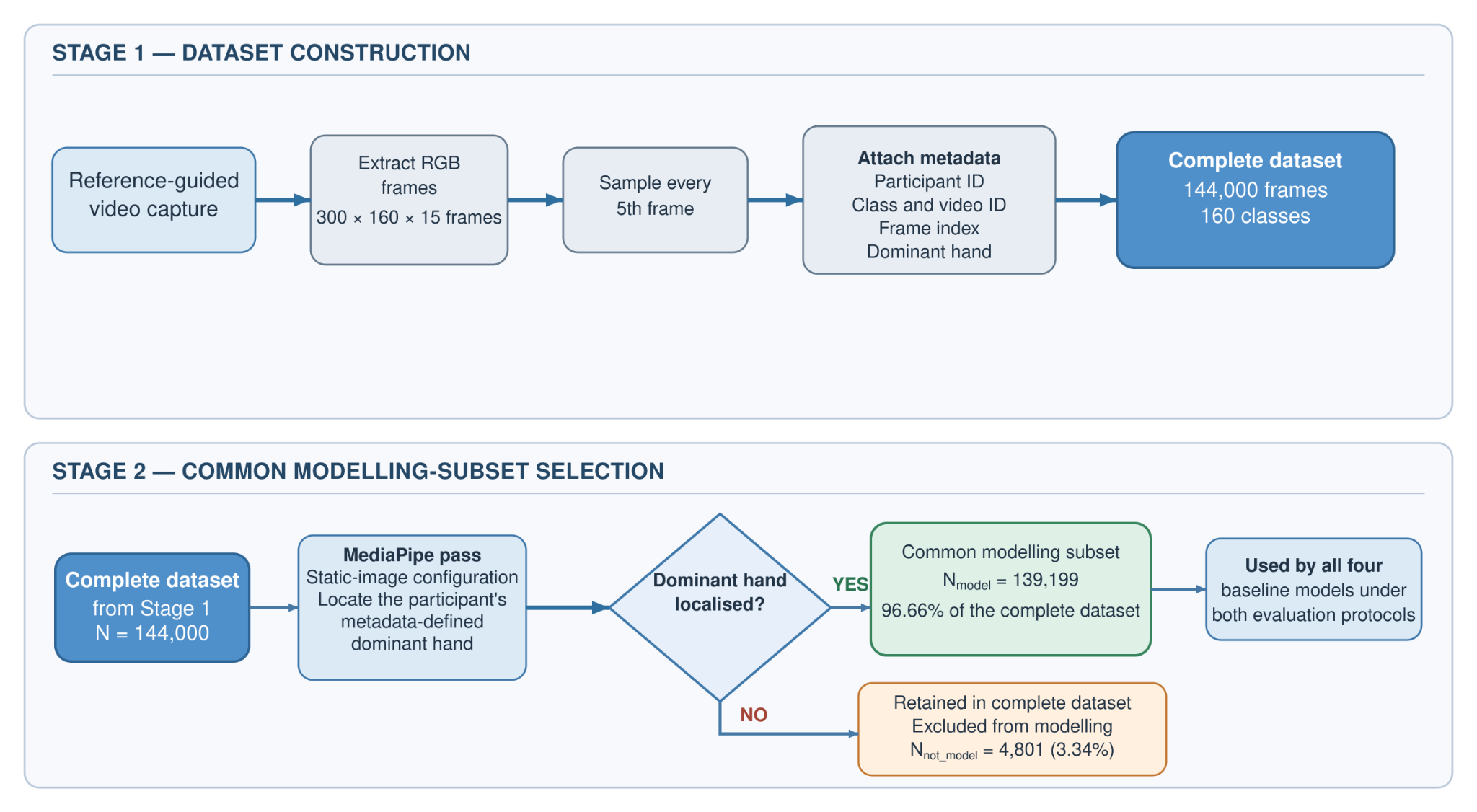}
\caption{Dataset construction and common modelling-subset selection pipeline.}
\label{fig:data-collection-flow}
\end{figure*}

Every class defined in
Section~\ref{subsec:hamnosys-handshape-inventory-and-class-definition}
was performed by each participant.
After the reference-guided verification described above, a 10-second
clip was recorded while the participant maintained the handshape and
slowly rotated the hand about two approximately orthogonal axes. This
efficiently introduced viewpoint, apparent overlap, and self-occlusion
variation without intentionally changing the class.
Each 10-second clip recorded at 30 frames/s yielded 300 RGB frames. To reduce temporal redundancy while retaining samples throughout the hand rotation, every fifth frame was selected, producing 60 images per clip. Participant, HamNoSys class, source-video, frame-index, and dominant-hand metadata were stored for each image, yielding 144,000 labelled images. MediaPipe Hands \cite{zhang2020mediapipe}, configured as described in Section~\ref{sec:experiments}, localised the metadata-defined dominant hand in 139,199 images. These images formed the common modelling subset for all four baselines, thereby preventing representation-dependent sample selection. The remaining 4,801 images were retained in the complete dataset. The pipeline is shown in Fig.~\ref{fig:data-collection-flow}.

\subsection{Dataset Organisation and Naming Convention}
\label{subsec:dataset-organisation}

The dataset is organised hierarchically to preserve the
provenance of every image and its correspondence with the chart-defined
handshape inventory. The root directory contains 15 participant folders,
labelled \texttt{s0001} to \texttt{s0015}. Each participant folder is
divided into the ten populated HamNoSys categories defined in
Table~\ref{tab:selection-inventory}. Within each category, images are
grouped by handshape subclass and chart variation.

A class folder follows the naming convention

\begin{center}
\texttt{\{subclass\}\_\{category\}\{variation\}},
\end{center}

where \texttt{\{subclass\}} identifies the relevant chart-column group,
\texttt{\{category\}} denotes the dataset-specific selection code, and
\texttt{\{variation\}} indexes the corresponding drawn hand model within
that combination. For example, \texttt{SFE\_OF2} denotes the second
\emph{Selected Fingers Extended} form in the \emph{One Finger} selection
category, whereas \texttt{DE\_F4} denotes the fourth
\emph{Derivation Example} associated with the \emph{Fist} category.

Individual image files retain their source-video identifier and frame
index using the convention

\begin{center}
\texttt{\{video\}\_frame\_\{frameindex\}.png}.
\end{center}

Thus, a path such as

\begin{center}
\texttt{s0001/OF/SFE\_OF2/v0000014\_frame\_00000.png}
\end{center}

identifies participant \texttt{s0001}, the \emph{One Finger} selection
category, the second \emph{Selected Fingers Extended} class, source video
\texttt{v0000014}, and extracted frame \texttt{00000}. Similarly,

\begin{center}
\texttt{s0001/F/DE\_F4/v0000111\_frame\_00000.png}
\end{center}

corresponds to the fourth derivation-example class under the
\emph{Fist} selection.

This naming scheme makes each image traceable to its participant,
selection category, chart-defined handshape class, source recording, and
frame position. A supplementary class-mapping file provides the complete
correspondence between the 160 dataset class identifiers and the
illustrated hand models in the official HamNoSys 4 Handshapes Chart.

\subsection{Dataset Characteristics and Class Distribution}
\label{subsec:dataset-characteristics}

The complete dataset and common modelling subset are summarised in
Table~\ref{tab:dataset-characteristics}. The 144,000 images correspond to 900 per
class and 60 per class per participant.

\begin{table}[!t]
\centering
\small
\caption{Principal characteristics of the HamNoSys handshape dataset.}
\label{tab:dataset-characteristics}
\renewcommand{\arraystretch}{1.08}
\begin{tabularx}{\linewidth}{@{}Xr@{}}
\toprule
Characteristic & Value \\
\midrule
Total images & 144,000 \\
Images used for modelling & 139,199 (96.66\%) \\
Images excluded from modelling & 4,801 (3.34\%) \\
Handshape classes & 160 \\
Participants & 15 \\
Image resolution & $640 \times 480$ pixels \\
Image modality & RGB \\
Mean images per class in total subset & 900 \\
Mean images per class in modelling subset & 869.99 \\
Minimum images in a class in modelling subset & 812 \\
Maximum images in a class in modelling subset & 900 \\

\bottomrule
\end{tabularx}
\end{table}

\begin{figure*}[!t]
\centering
\includegraphics[width=\linewidth]
  {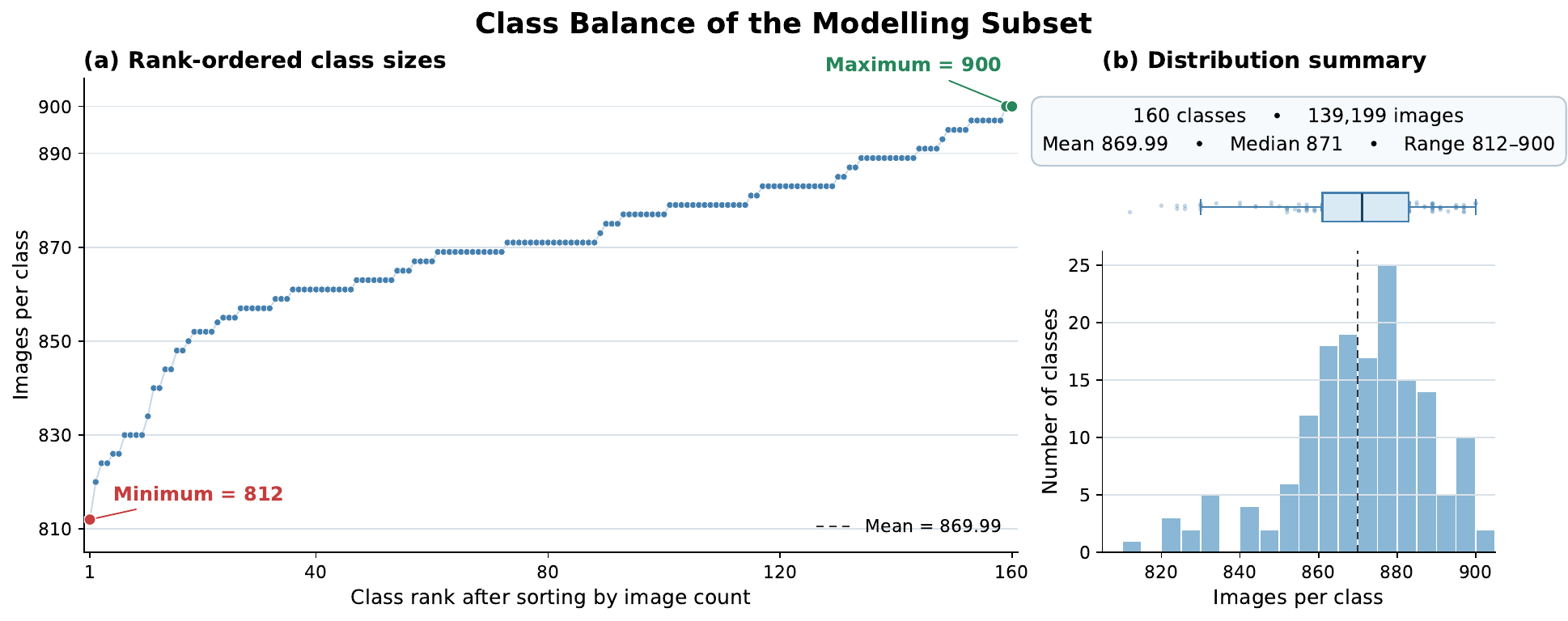}
\caption{Class balance of the 139,199-image modelling subset.
(a) Rank-ordered image counts for the 160 handshape classes, with
the minimum and maximum class sizes highlighted. The dashed line
indicates the mean class size. (b) Boxplot and histogram summarizing
the distribution of per-class image counts. Class sizes range from
812 to 900 images, with a mean of 869.99 images per class.}
\label{fig:class-distribution}
\end{figure*}

The modelling subset remains approximately uniform
(Fig.~\ref{fig:class-distribution}), ranging from 812 images for
\texttt{DE\_F3} to 900 for \texttt{FTR\_TFO1} and
\texttt{SFB\_TFS3}.

\begin{figure*}[!t]
\centering
\setlength{\tabcolsep}{2pt}
\begin{tabular}{@{}cccc@{}}
\includegraphics[width=0.235\linewidth]{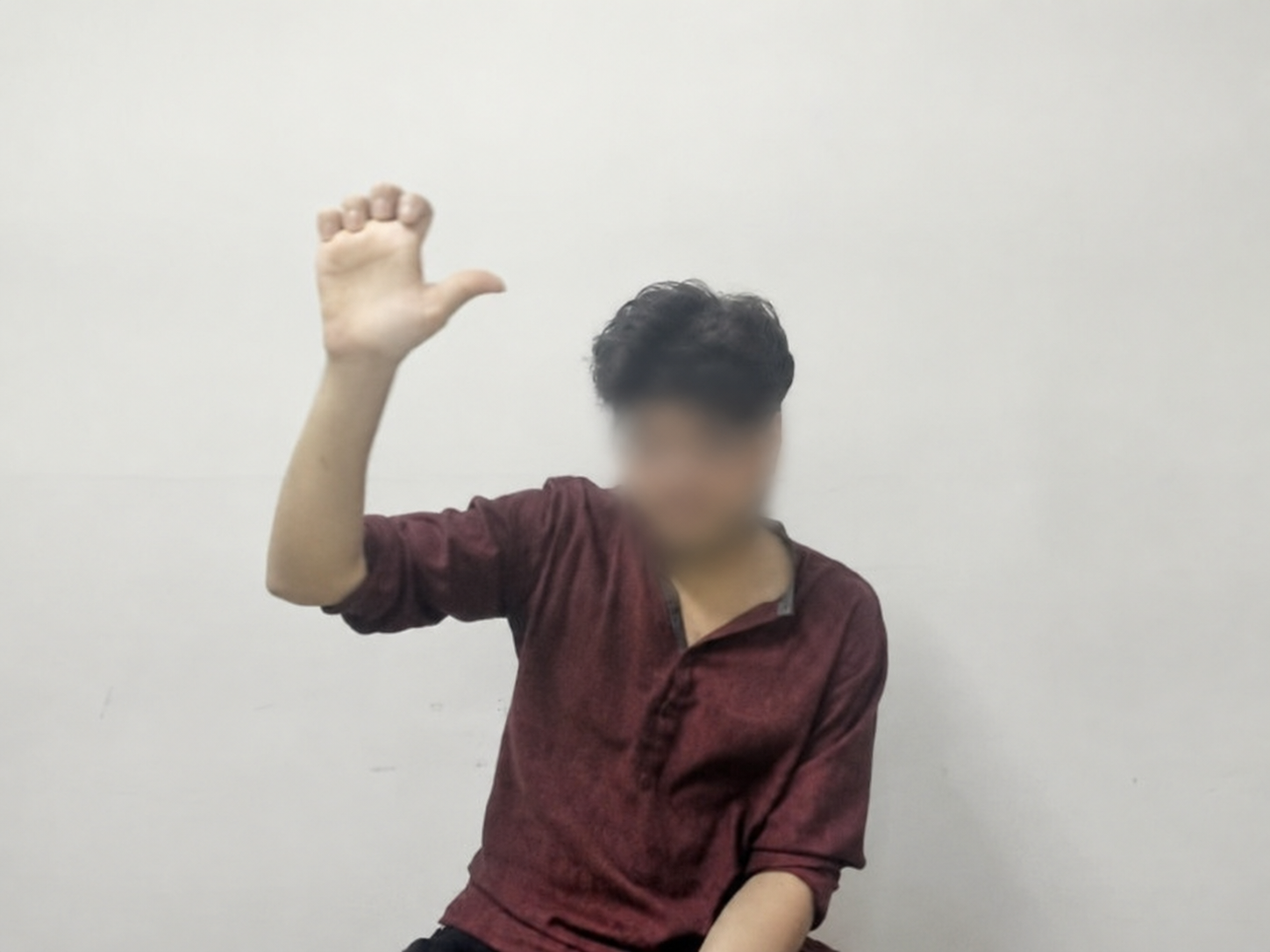} &
\includegraphics[width=0.235\linewidth]{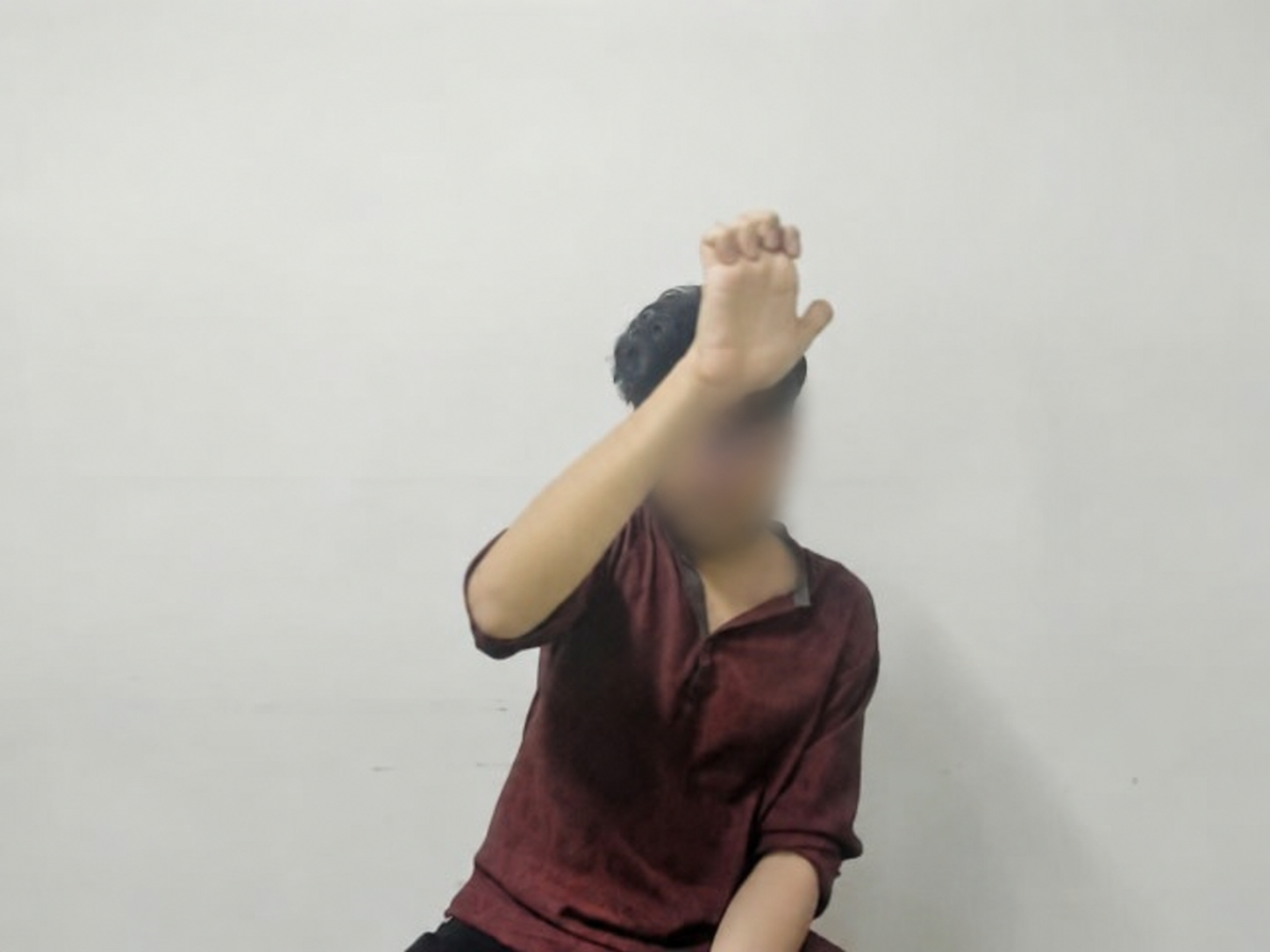} &
\includegraphics[width=0.235\linewidth]{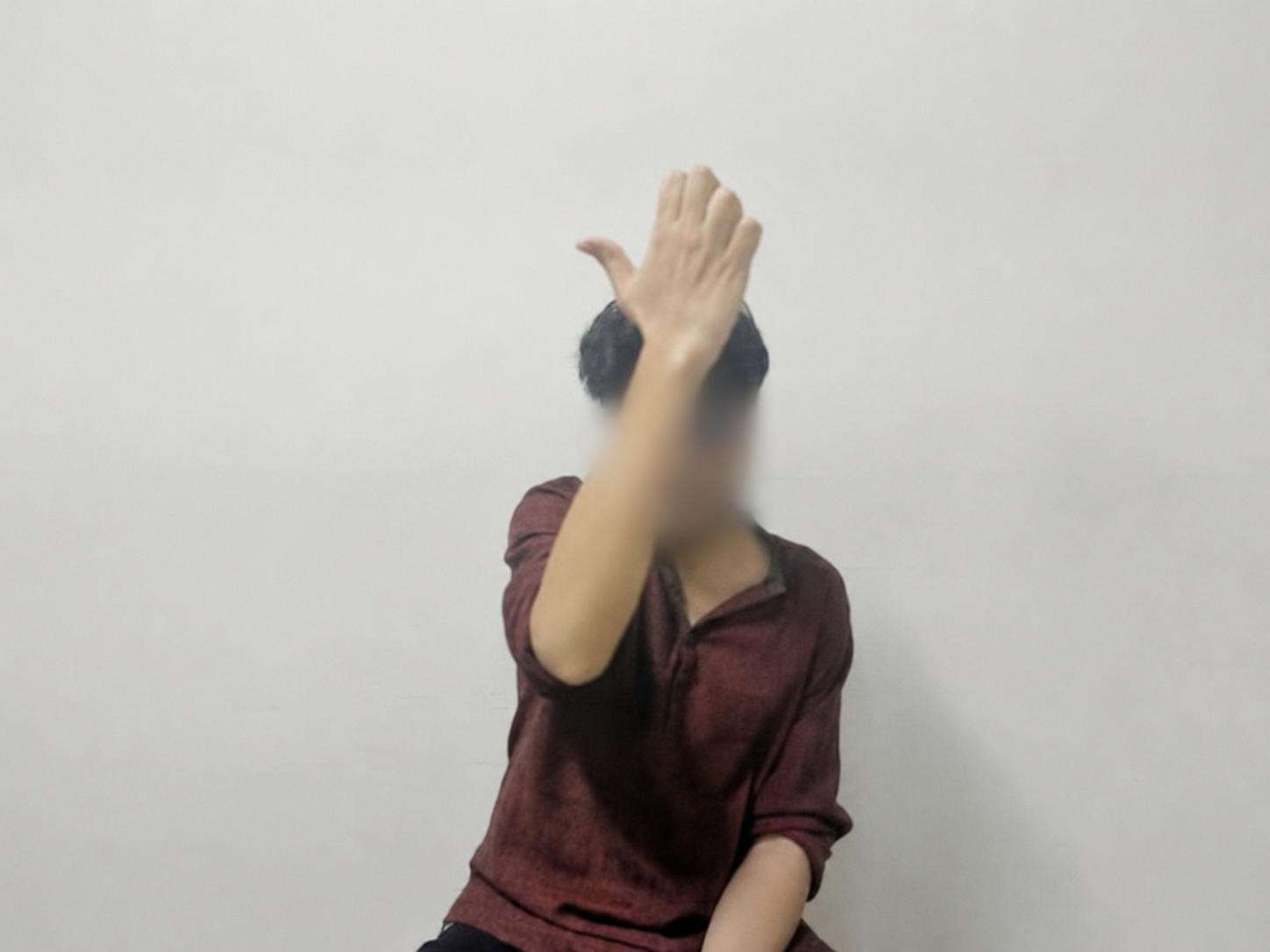} &
\includegraphics[width=0.235\linewidth]{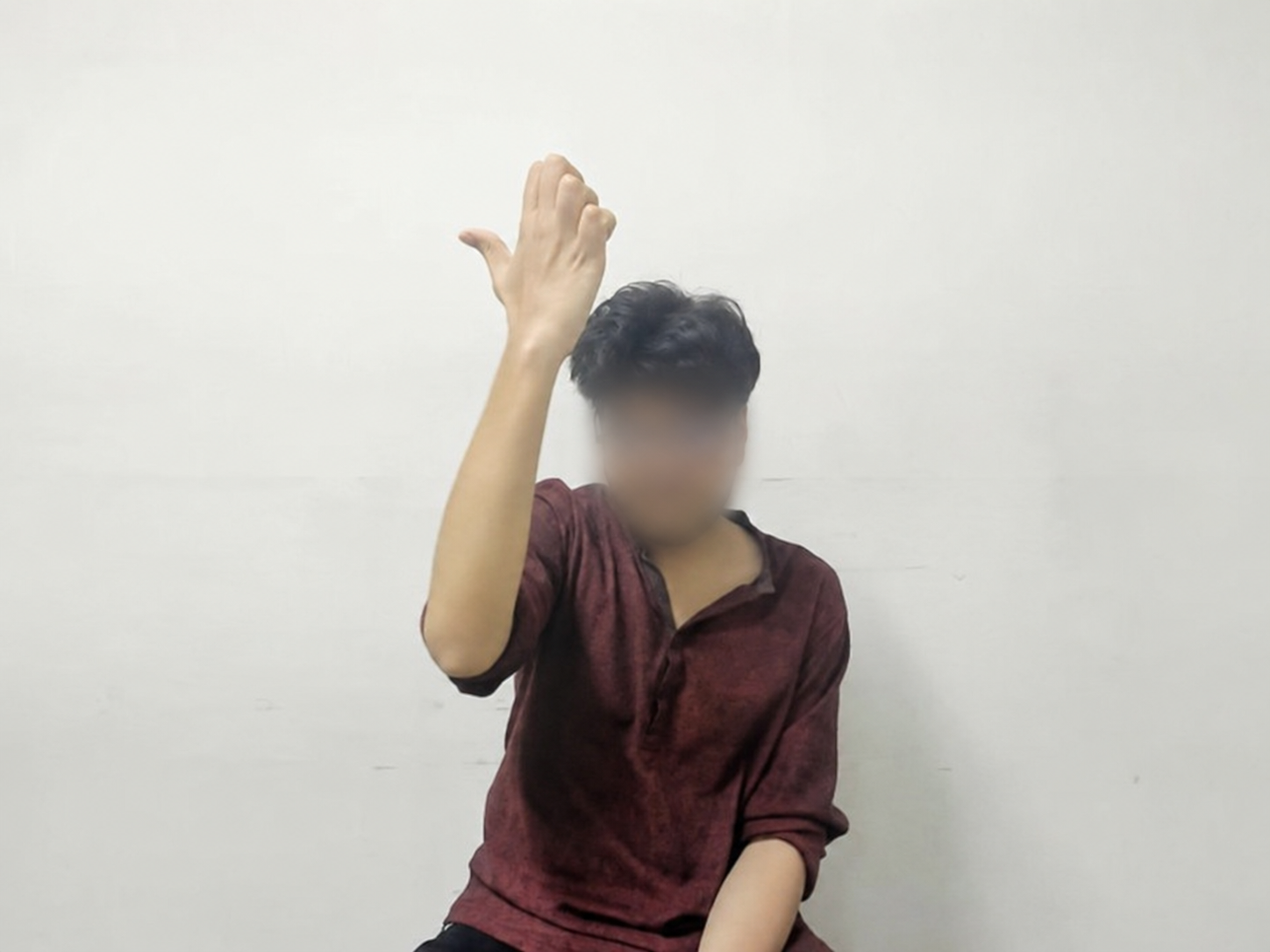} \\[2pt]
\includegraphics[width=0.235\linewidth]{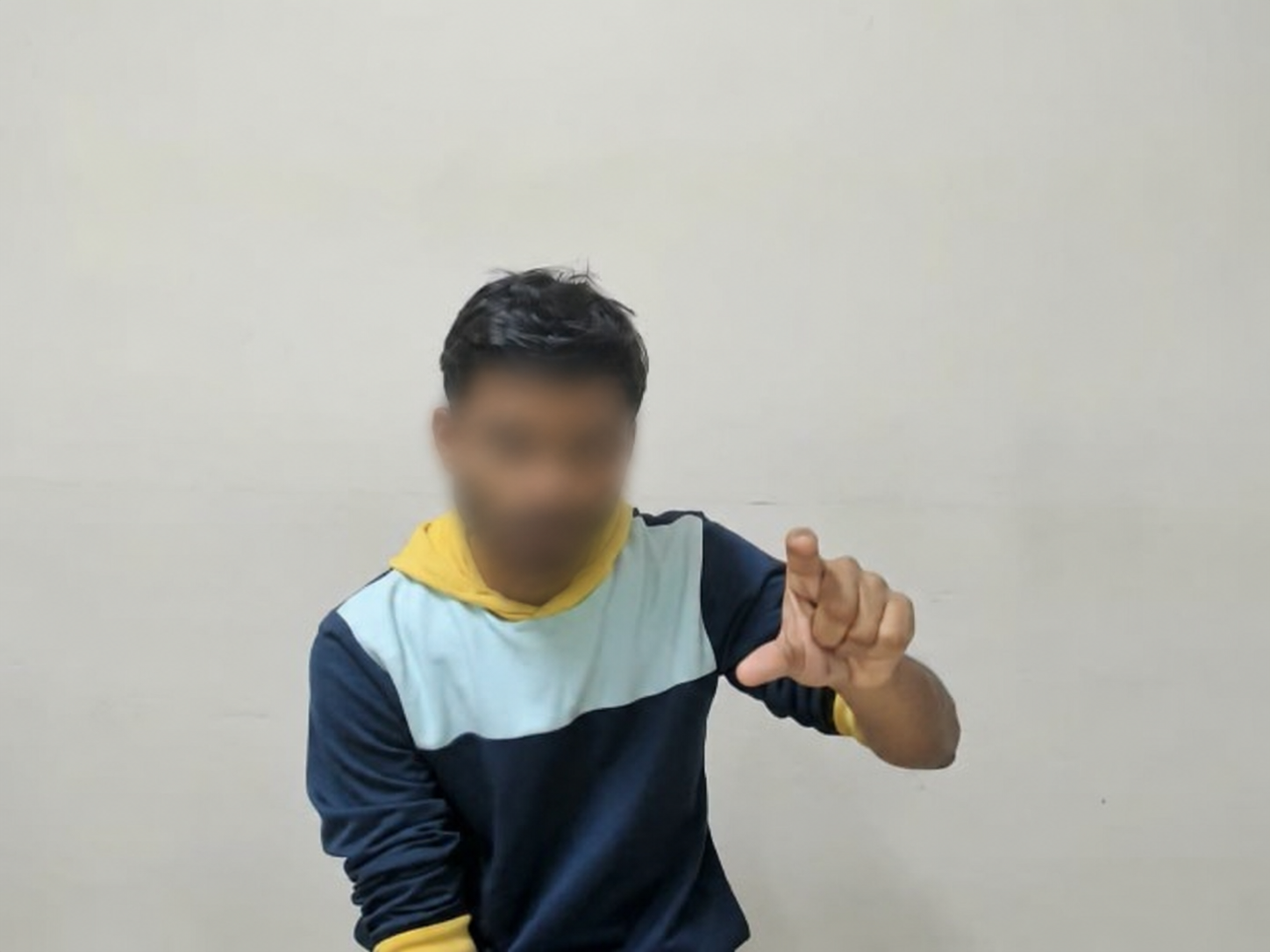} &
\includegraphics[width=0.235\linewidth]{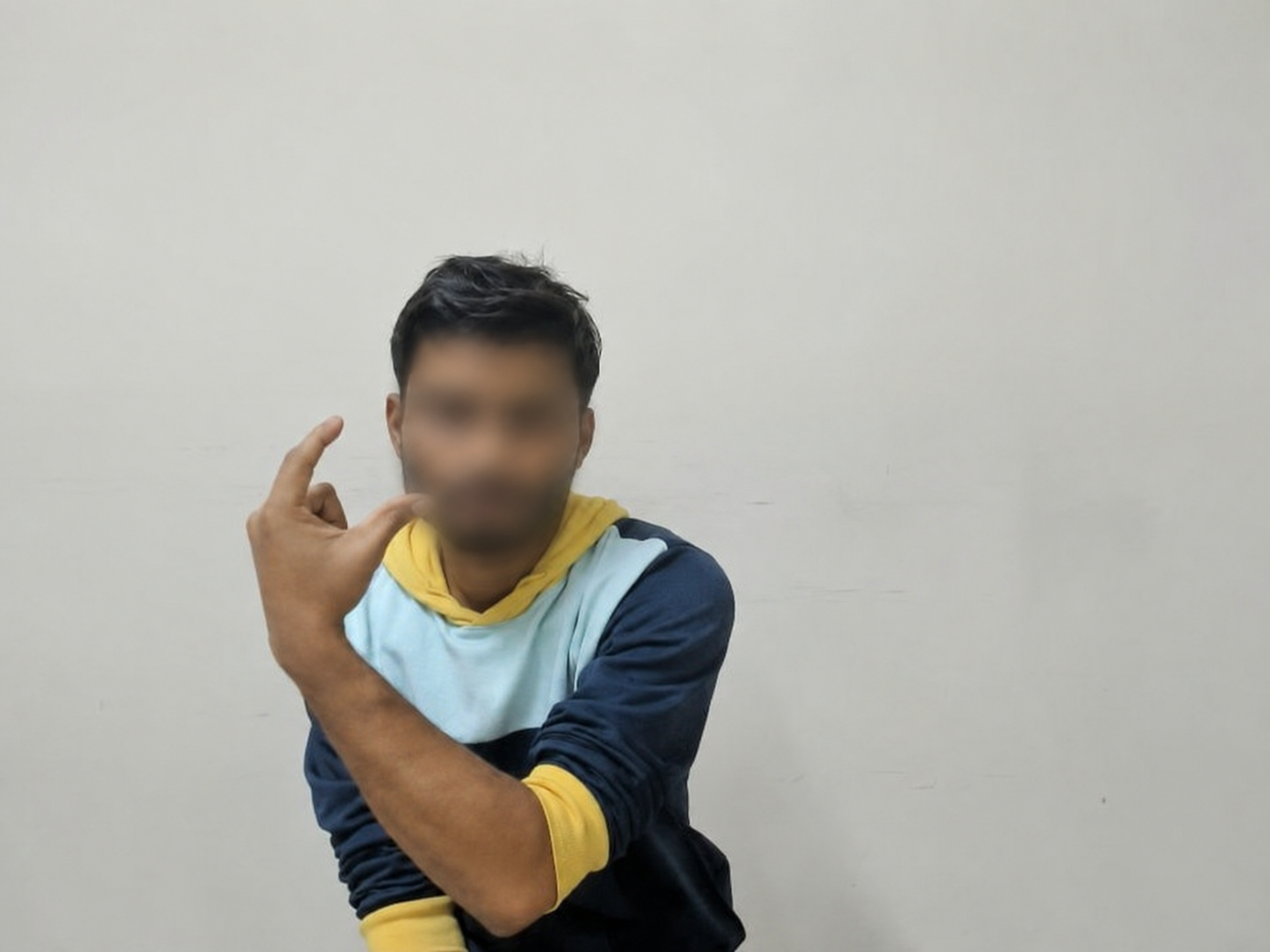} &
\includegraphics[width=0.235\linewidth]{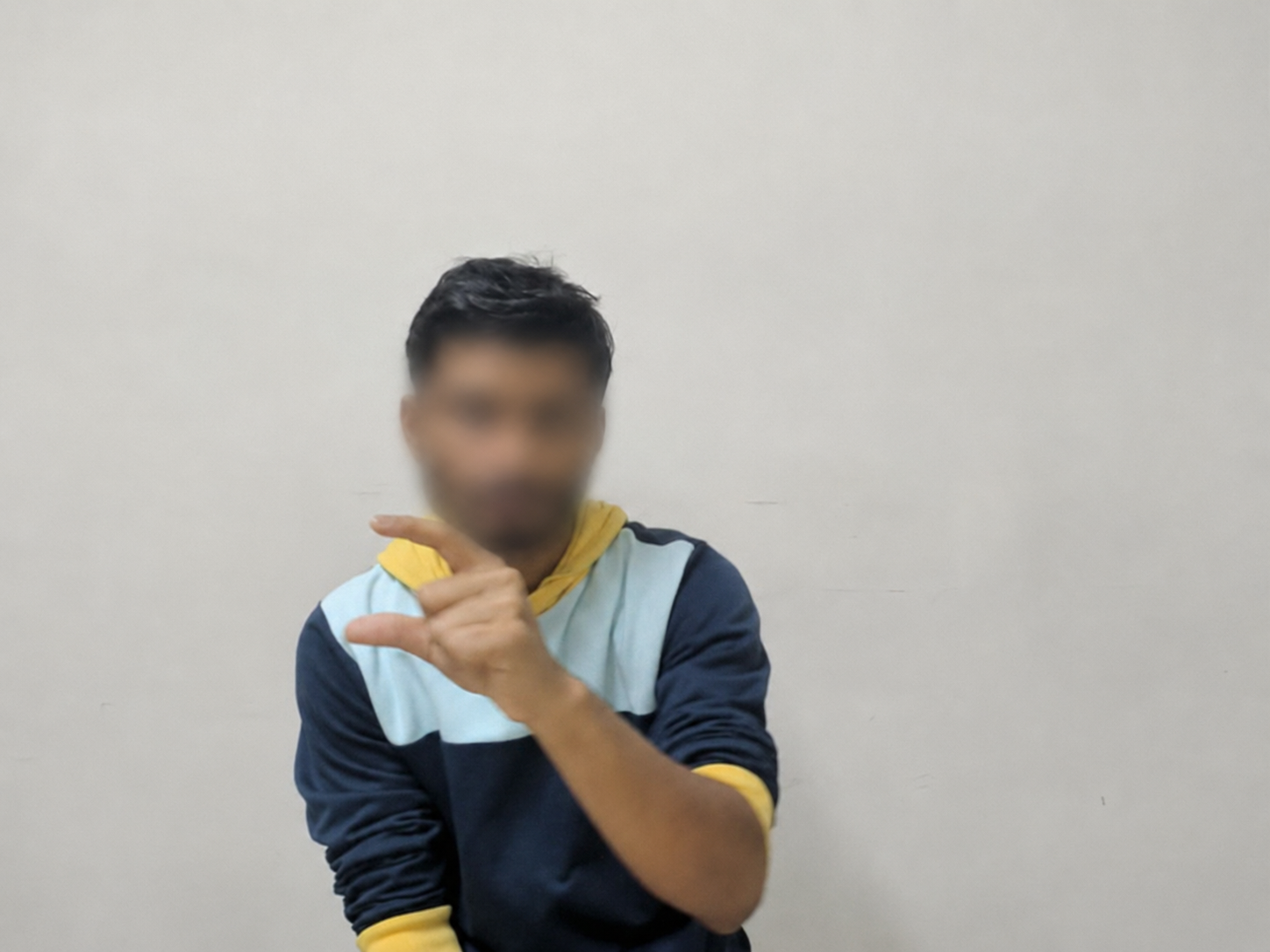} &
\includegraphics[width=0.235\linewidth]{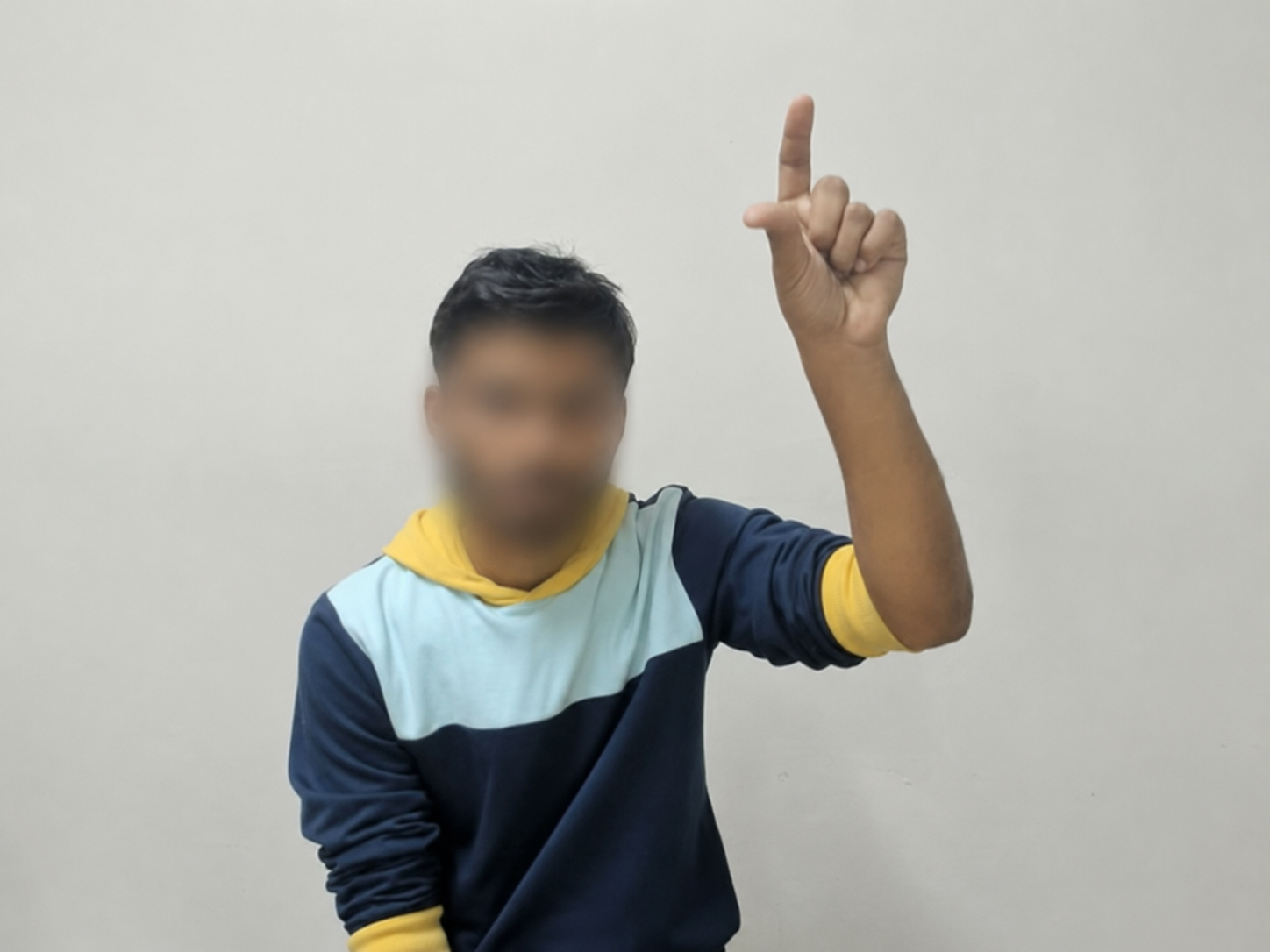}
\end{tabular}
\caption{Representative dataset frames from two participants. Each row
contains four frames from one participant, illustrating
variation in viewpoint, hand position, articulation, and appearance.
Faces are blurred to protect participant identity.}
\label{fig:representative-samples}
\end{figure*}

Within-class variation in viewpoint, apparent orientation, hand position,
and self-occlusion was introduced by the video-based acquisition
procedure. Figure~\ref{fig:representative-samples} presents four frames from each of two representative recordings,
with one participant , enacting different variations of a handshape, shown per row.

\subsection{Evaluation Protocols and Data Splits}
\label{subsec:evaluation-protocols}

Two complementary protocols are defined over the 139,199-image modelling
subset. A class-stratified frame-level split is used to provide a subject-dependent
reference comparable with conventional image-classification benchmarks,
including datasets for which signer identities are unavailable. Because
participant identity was preserved during collection, the stricter
assessment of generalisation to unseen participants is provided by a
15-fold leave-one-subject-out (LOSO) protocol.

\subsubsection{Subject-Dependent Protocol}

For the subject-dependent evaluation, images were partitioned at frame
level using a class-stratified random 70:15:15 split. Temporal redundancy
was reduced before this split by retaining every fifth frame, as
described in Section~\ref{subsec:acquisition-system}. The resulting counts
are reported in Table~\ref{tab:subject-dependent-split}. Because all
participants may occur in every partition, recognition under
seen-participant conditions is measured by this protocol;
unseen-participant generalisation is assessed separately by LOSO.

\begin{table}[!t]
\centering
\small
\caption{Subject-dependent frame-level split of the 139,199-image
modelling subset.}
\label{tab:subject-dependent-split}
\renewcommand{\arraystretch}{1.08}
\begin{tabularx}{\linewidth}{@{}Xrrr@{}}
\toprule
Partition & Images & Percentage & Approx. per class \\
\midrule
Training   & 97,370  & 69.95\% & 609 \\
Validation & 20,806  & 14.95\% & 130 \\
Test       & 21,023  & 15.10\% & 131 \\
\midrule
\textbf{Total} & \textbf{139,199} & \textbf{100\%} & \textbf{870} \\
\bottomrule
\end{tabularx}
\end{table}

\subsubsection{Subject-Independent Protocol (LOSO)}

Subject-independent performance was evaluated using 15-fold leave-one-subject-out cross-validation, as summarised in Table~\ref{tab:loso-protocol}. In each fold, one participant was reserved exclusively for testing, while images from the remaining 14 participants were divided into class-stratified training and validation partitions using an 85:15 ratio. Identical partitions were used for all four baselines, and no data from the held-out participant were used for model fitting, validation, checkpoint selection, or other training-stage decisions. Each participant served as the test subject once, and performance was reported as the mean and sample standard deviation across the 15 folds.

\begin{table}[!t]
\centering
\small
\caption{Structure of each leave-one-subject-out fold. Exact image counts
vary with the number of usable images contributed by the held-out
subject.}
\label{tab:loso-protocol}
\renewcommand{\arraystretch}{1.08}
\begin{tabularx}{\linewidth}{@{}>{\raggedright\arraybackslash}p{3.1cm}>{\raggedright\arraybackslash}X@{}}
\toprule
Component & Definition \\
\midrule
Test set &
All usable images from one held-out subject \\

Development set &
All usable images from the remaining 14 subjects \\

Training partition &
85\% of the development images, selected using
class-stratified sampling \\

Validation partition &
15\% of the development images, selected using the same
class-stratified split \\

Number of folds &
15, with each subject held out exactly once \\

Subject overlap &
None between the test subject and the training or validation partitions \\
\bottomrule
\end{tabularx}
\end{table}

\section{Experiments}
\label{sec:experiments}

\begin{figure*}[!t]
\centering
\includegraphics[width=\linewidth]
{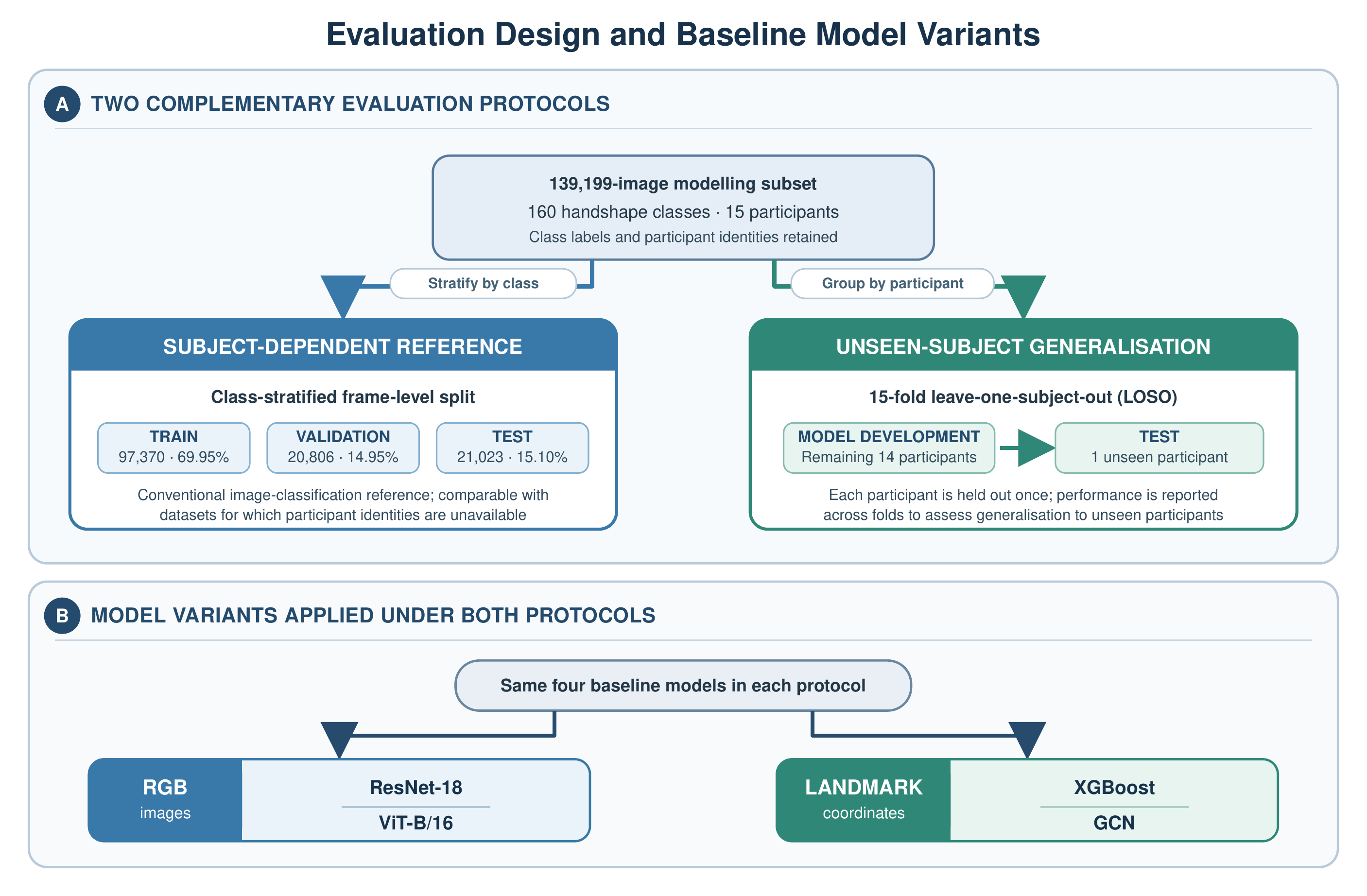}
\caption{Overview of the evaluation design and baseline model variants.}
\label{fig:evaluation-design}
\end{figure*}

Baseline recognition performance is established for the proposed dataset
under the subject-dependent and leave-one-subject-out
protocols defined in Section~\ref{subsec:evaluation-protocols}. Four
models were evaluated: two appearance-based models operating on RGB hand
crops and two landmark-based models operating on MediaPipe hand
keypoints. The comparisons are intended to characterise the benchmark
rather than propose a new recognition architecture.Figure~\ref{fig:evaluation-design} presents the two complementary
evaluation protocols and the model variants used under each protocol.

External context is provided by two datasets. LSWH100
\cite{loboneto2024} contains 144,000 synthetic images in 100
SignWriting-derived classes with predefined training, validation, and
test partitions; model output layers were changed to 100 classes and
results computed on the predefined test set. ASL Fingerspelling Dataset A
\cite{pugeault2011} contains 65,774 real RGB observations from five users
covering 24 static ASL letters (excluding dynamic J and Z). Output layers
were changed to 24 classes. A class-stratified 70:15:15
subject-dependent split and five-fold LOSO were used, with 15\% of each
remaining-user development set reserved for validation where required.

These are matched-model references, not direct rankings of dataset
quality: image origin, class definition and count, participant
composition, and acquisition protocol differ across datasets.

\subsection{Training Configurations and Baseline Models}
\label{subsec:experimental-models}

\paragraph{RGB Input based models.}
For the appearance-based models, MediaPipe Hands
\cite{zhang2020mediapipe} was applied in static-image mode to identify
the participant's metadata-defined dominant hand. The hand bounding box
was expanded by 15 pixels on each side, subject to the image boundaries,
and the resulting crop was resized to $224\times224$ pixels.

For ImageNet-pretrained ResNet-18 \cite{he2016}, the classifier was
replaced by a 160-class linear head and only the final residual stage and
new head were fine-tuned. For ImageNet-pretrained ViT-B/16
\cite{dosovitskiy2021}, the head was replaced by dropout (0.3) and a
160-class linear layer; the final two transformer blocks, encoder
normalisation, and head were fine-tuned. Horizontal flipping, rotation up
to $15^{\circ}$, colour jitter, and random erasing were used for
ResNet-18 augmentation. Affine and perspective transformations and
stronger photometric augmentation were additionally used for ViT-B/16.
Evaluation images were resized and ImageNet-normalised.

\paragraph{Landmark Input based models.}
MediaPipe Hands was also used to extract 21 dominant-hand landmarks from
each image, which were preprocessed as in \cite{sarkar24}. Detected left hands were mirrored to a common canonical
orientation. The coordinates were translated so that the wrist lay at
the origin, rotated into a hand-centred coordinate frame, and scaled by
the maximum landmark distance. A joint-angle feature, normalised by
$\pi$, was computed for the 15 internal finger joints; the wrist and
fingertips were assigned an angle of zero. Each node was therefore
represented by
\[
(x,y,z,\theta),
\]
giving a $21\times4$ feature matrix.

The anatomical hand connections were used as edges by the graph
convolutional network, with self-connections introduced during graph
convolution. Five graph-convolutional layers with output dimensions
$[512,480,448,416,352]$, GELU activations, residual connections, batch
normalisation, dropout of 0.1, global mean pooling, and a 160-class
linear output layer were included.

An 84-dimensional vector formed from each landmark's $x$, $y$, and $z$
coordinates and joint angle was used by the XGBoost baseline
\cite{chen2016}:
\[
21\times(x,y,z,\theta).
\]
For XGBoost, 1,200 boosting rounds were used with maximum depth 10,
minimum child weight 3, learning rate 0.02180, gamma 0.01248,
subsample ratio 0.6762, column-sampling ratio 0.7482,
$\ell_2$ regularisation 0.01875, $\ell_1$ regularisation 0.09311,
and histogram-based tree construction.

\begin{table*}[!t]
\centering
\small
\caption{Training configuration of the neural baseline models.}
\label{tab:training-config}
\renewcommand{\arraystretch}{1.08}
\begin{tabularx}{\linewidth}{@{}Xccc@{}}
\toprule
Setting & ResNet-18 & ViT-B/16 & GCN \\
\midrule
Pretraining
  & ImageNet & ImageNet & None \\
Optimizer
  & Adam & AdamW & AdamW \\
Learning rate
  & $5\times10^{-5}$
  & $1\times10^{-4}$
  & $6.451\times10^{-4}$ \\
Weight decay
  & $1\times10^{-4}$
  & $5\times10^{-4}$
  & $3.285\times10^{-5}$ \\
Batch size
  & 64 & 32 & 128 \\
Maximum epochs
  & 60 & 60 & 150 \\
Early-stopping patience
  & 8 & 6 & 10 \\
Learning-rate schedule
  & ReduceLROnPlateau
  & CosineAnnealingLR
  & ReduceLROnPlateau \\
Label smoothing
  & 0.1 & 0.1 & None \\
Input
  & $224\times224$ RGB
  & $224\times224$ RGB
  & $21\times4$ graph \\
\bottomrule
\end{tabularx}
\end{table*}

\subsection{Evaluation Measures}
\label{subsec:evaluation-measures}

The subject-dependent results for the proposed dataset and the
matched-model results for LSWH100 and ASL Fingerspelling Dataset A are
reported in Table~\ref{tab:subject-dependent-results}. Because each
partition may contain frames from all participants, this protocol
measures recognition under participant overlap and provides a
seen-participant reference. It is not interpreted as an estimate of
generalisation to unseen participants.
\subsection{Results}
\label{subsec:results}

\subsubsection{Subject-Dependent Performance and External Reference}

The subject-dependent results for the proposed dataset and the
matched-model LSWH100 and ASL Fingerspelling Dataset A references are
reported in Table~\ref{tab:subject-dependent-results}.
Because frames from all participants may occur in each partition, these
values provide a seen-signer reference rather than an estimate of
performance on an unseen signer.

\begin{table*}[!t]
\centering
\small
\caption{Within-dataset test performance on the proposed HamNoSys
dataset and ASL Fingerspelling Dataset A under subject-dependent
evaluation, and on LSWH100 using its predefined split (\%).}
\label{tab:subject-dependent-results}
\renewcommand{\arraystretch}{1.08}
\setlength{\tabcolsep}{2.5pt}

\begin{tabular*}{\linewidth}
{@{\extracolsep{\fill}}lccccccccc@{}}
\toprule
& \multicolumn{3}{c}{Accuracy}
& \multicolumn{3}{c}{Weighted average}
& \multicolumn{3}{c}{Macro average} \\
\cmidrule(lr){2-4}
\cmidrule(lr){5-7}
\cmidrule(l){8-10}
Model
& @1 & @3 & @5
& P & R & $F_1$
& P & R & $F_1$ \\
\midrule
\multicolumn{10}{@{}l}{\textit{Proposed HamNoSys dataset (160 classes)}} \\
ResNet-18
& 84.72 & 94.62 & 96.78
& 85.07 & 84.72 & 84.74
& 85.09 & 84.72 & 84.75 \\

ViT-B/16
& 86.20 & 95.99 & 97.79
& 86.50 & 86.20 & 86.19
& 86.51 & 86.19 & 86.20 \\

GCN
& 72.44 & 88.44 & 92.74
& 72.62 & 72.44 & 72.38
& 72.60 & 72.41 & 72.36 \\

XGBoost
& 69.57 & 85.23 & 90.06
& 69.76 & 69.57 & 69.55
& 69.74 & 69.54 & 69.53 \\
\addlinespace
\multicolumn{10}{@{}l}{\textit{LSWH100 (100 classes)}} \\
ResNet-18
& 86.70 & 97.17 & 98.67
& 87.11 & 86.70 & 86.69
& 87.11 & 86.70 & 86.69 \\

ViT-B/16
& 81.55 & 95.85 & 98.00
& 82.70 & 81.55 & 81.53
& 82.70 & 81.55 & 81.53 \\

GCN
& 79.54 & 94.82 & 96.79
& 80.11 & 79.54 & 79.52
& 79.80 & 79.54 & 79.36 \\

XGBoost
& 72.26 & 90.50 & 94.76
& 72.96 & 72.26 & 72.29
& 72.71 & 72.23 & 72.14 \\
\addlinespace
\multicolumn{10}{@{}l}{\textit{ASL Fingerspelling Dataset A (24 classes)}} \\
ResNet-18
& 99.94 & 100.00 & 100.00
& 99.94 & 99.94 & 99.94
& 99.94 & 99.94 & 99.94 \\

ViT-B/16
& 99.72 & 100.00 & 100.00
& 99.72 & 99.72 & 99.72
& 99.72 & 99.71 & 99.71 \\

GCN
& 97.21 & 99.09 & 99.35
& 97.25 & 97.21 & 97.21
& 97.17 & 96.98 & 97.05 \\

XGBoost
& 97.66 & 99.17 & 99.45
& 97.69 & 97.66 & 97.67
& 97.41 & 97.52 & 97.46 \\
\bottomrule
\end{tabular*}

\vspace{0.25em}
\footnotesize P = precision; R = recall. The external datasets have
different class inventories and acquisition conditions; their results
therefore provide context rather than direct estimates of relative
dataset quality.
\end{table*}

On the proposed 160-class dataset, the highest top-1, top-3, and top-5
accuracies were achieved by ViT-B/16, at 86.20\%, 95.99\%, and 97.79\%,
respectively. ResNet-18 achieved a comparable top-1 accuracy of 84.72\%.
The higher top-1 accuracies of the two RGB-based models relative to the
landmark-based baselines are consistent with fine-grained distinctions
benefiting from appearance information that is not completely retained
by the 21-point landmark representation. Among the landmark-based
models, GCN exceeded XGBoost by 2.87 percentage points in top-1
accuracy, suggesting an advantage from explicitly representing the
anatomical connectivity of the hand.

For every model, the weighted and macro scores were closely aligned.
This agreement is consistent with the approximately balanced class
distribution and indicates that the aggregate results were not
dominated by a small number of larger classes.

The highest LSWH100 top-1 accuracy was obtained by ResNet-18 at 86.70\%.
Relative to the proposed dataset, the change in top-1 accuracy ranged
from $-4.65$ to $+7.10$ percentage points and differed across models.
This change in model ordering, together with the different image
origins and 100- versus 160-class inventories, prevents a direct ranking
of dataset quality from these within-dataset results.

\begin{figure*}[!t]
\centering
\includegraphics[width=\linewidth]
{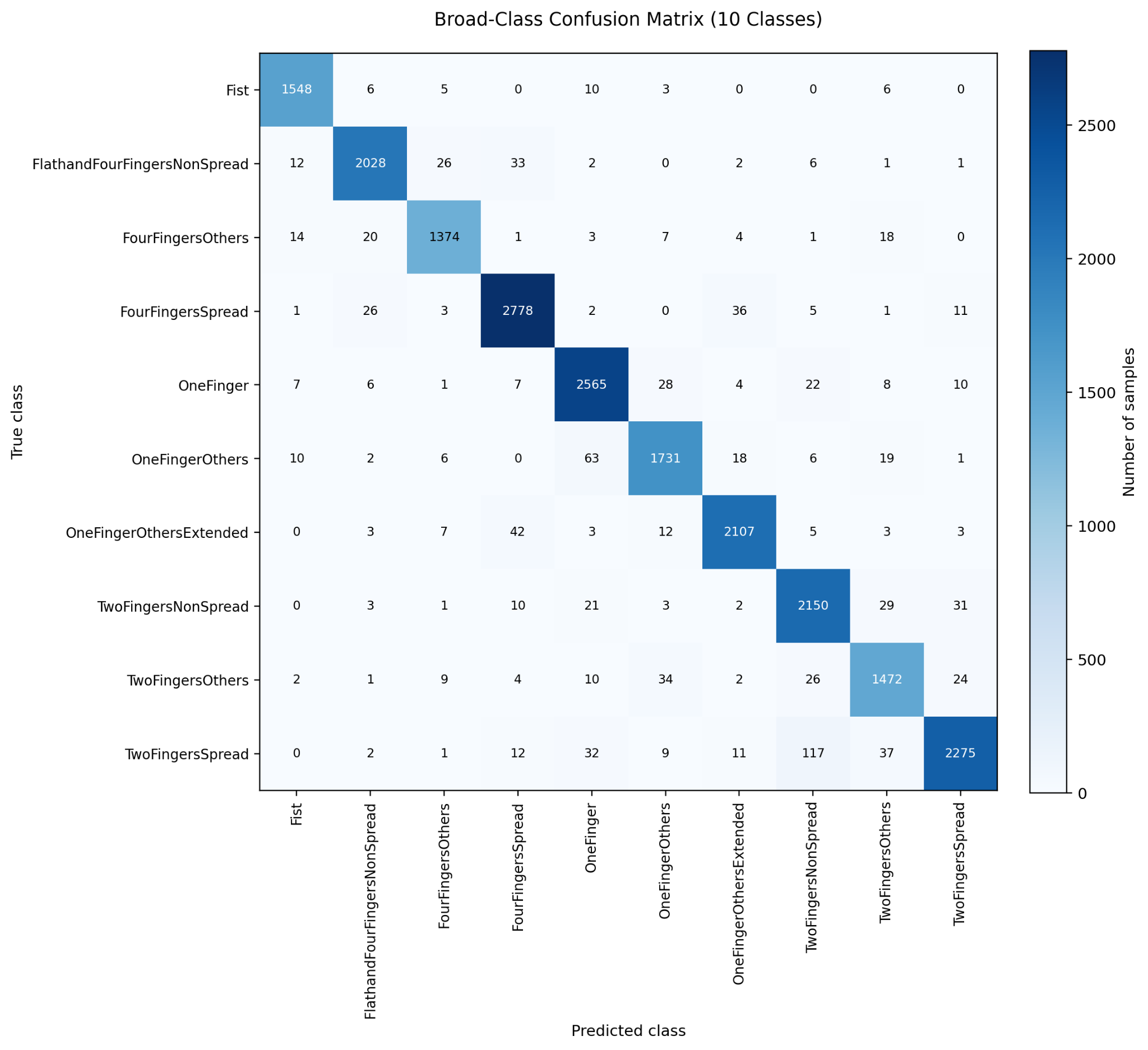}
\caption{Broad-category confusion matrix for ViT-B/16 under
subject-dependent evaluation. Rows denote true categories and columns
denote predicted categories; cell values are test-sample counts.}
\label{fig:vit-broad-confusion-matrix}
\end{figure*}

Top-1 accuracy above 97\% was obtained by all four models on ASL
Fingerspelling Dataset A. These near-ceiling participant-overlapping
results must be interpreted in relation to its smaller 24-class
alphabetic inventory and different acquisition conditions. The
external evaluations therefore establish matched-model reference
points rather than direct measures of the relative quality of the
three datasets.
Because ViT-B/16 achieved the strongest subject-dependent performance
on the proposed dataset, it was selected for the subsequent confusion
analysis. The broad-category confusion matrix in
Figure~\ref{fig:vit-broad-confusion-matrix} is strongly concentrated
along the diagonal, indicating that the ten broad handshape categories
were generally separated successfully. The remaining off-diagonal
predictions occurred primarily between related categories and motivated
a finer target-level analysis.
 
\begin{figure*}[!t]
\centering
\includegraphics[width=\linewidth,keepaspectratio]
{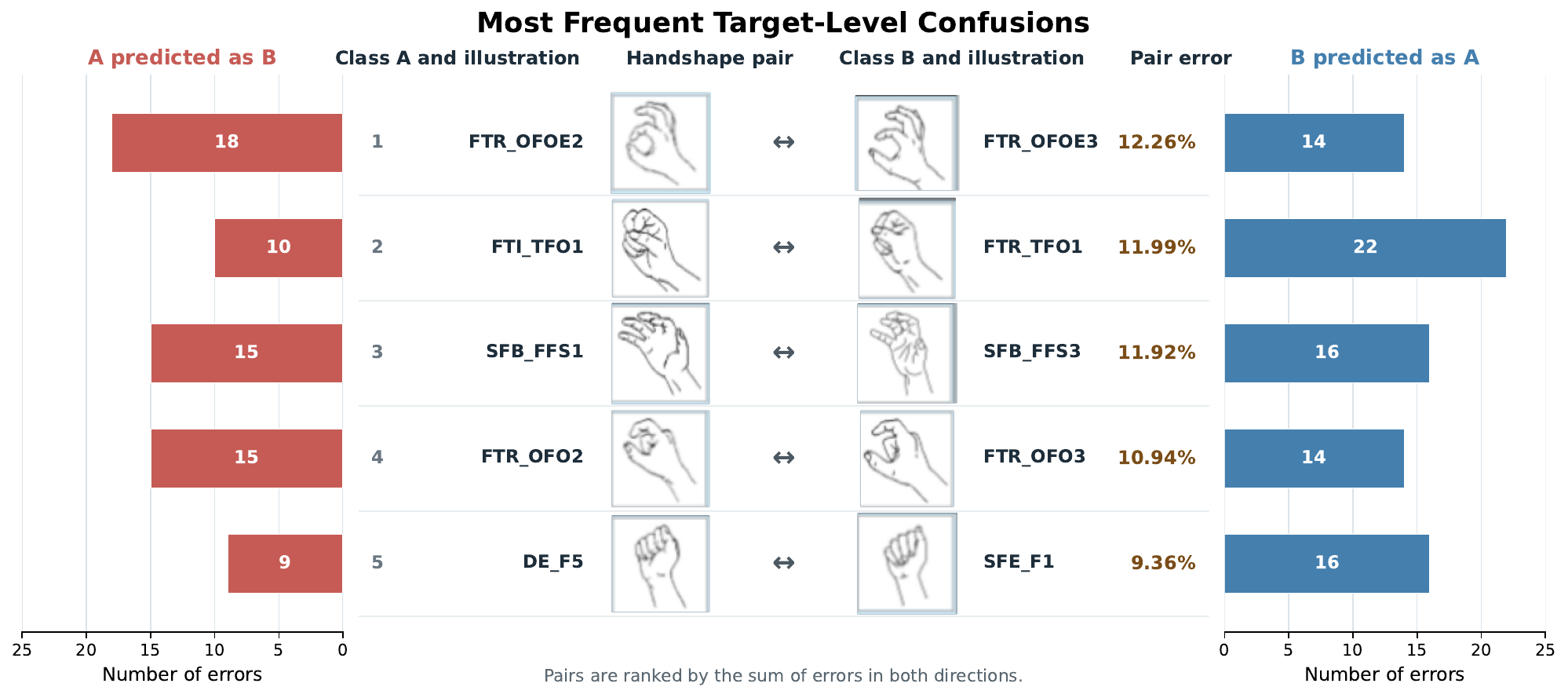}
\caption{Five most frequently confused target-level handshape pairs for
ViT-B/16 under subject-dependent evaluation. HamNoSys chart illustrations
are displayed alongside the corresponding class codes. The left and right
bars report directional misclassification counts, while the pair error
rate represents the total bidirectional errors relative to the combined
support of the two classes.}
\label{fig:top-five-confused-handshapes}
\end{figure*}

As shown in Figure~\ref{fig:hamnosys-chart}, several target classes
differ only in subtle properties such as finger selection, bending,
thumb position, or contact. The five target-level pairs with the
largest bidirectional confusion counts are shown in
Figure~\ref{fig:top-five-confused-handshapes}. For each pair, the pair
error rate was calculated as the total number of errors in both
directions divided by the combined support of the two classes.

\subsubsection{Leave-One-Subject-Out Performance}

The LOSO results for the proposed dataset and ASL Fingerspelling
Dataset A are reported in Table~\ref{tab:loso-results}. LOSO evaluation
could not be conducted on LSWH100 because signer-identity information
is not provided with that dataset.

On the proposed dataset, numerically close mean top-1 accuracies were
obtained by ResNet-18 and ViT-B/16, at 45.38\% and 45.22\%,
respectively. Although its top-1 accuracy was lower, GCN achieved the
highest top-3 and top-5 accuracies, at 69.23\% and 78.58\%. This result
indicates that the landmark graph frequently retained the correct class
among its leading predictions even when the top-ranked prediction was
incorrect.

Relative to the participant-overlapping evaluation, the ResNet-18 and
ViT-B/16 top-1 accuracies decreased by 39.34 and 40.98 percentage
points, respectively. The fold standard deviations also demonstrate
substantial variation among held-out participants. These results
identify unseen-participant generalisation as the principal challenge
of the proposed 160-class benchmark rather than indicating a failure of
within-participant handshape recognition.

On ASL Fingerspelling Dataset A, mean LOSO top-1 accuracy ranged from
82.20\% for ViT-B/16 to 87.40\% for ResNet-18. These values are not
directly comparable with those of the proposed dataset because ASL
Dataset A contains only 24 classes and was collected under different
conditions. In contrast, the proposed inventory contains 160
fine-grained classes, including several visually similar handshapes
that differ in limited articulatory properties. The confusion pairs in
Figure~\ref{fig:top-five-confused-handshapes} illustrate this
fine-grained separation problem. Improved representations of local
finger articulation and greater robustness to inter-participant
variation are therefore important directions for further modelling.

\begin{figure*}[h]
\centering
\includegraphics[width=0.92\linewidth]
  {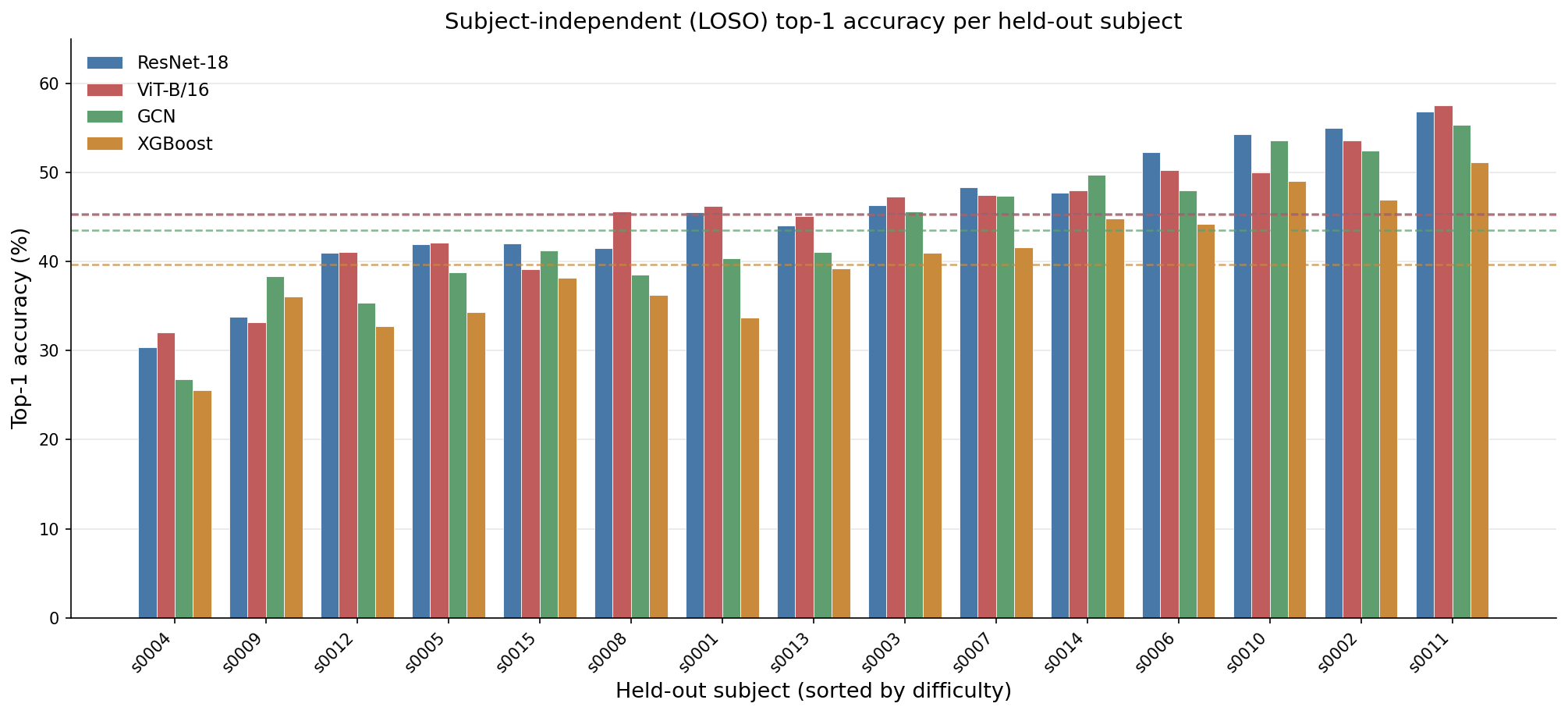}
\caption{Top-1 accuracy for each held-out
signer under LOSO evaluation. Dashed lines indicate the respective
across-signer means.}
\label{fig:loso-results}
\end{figure*}

Per-participant top-1 accuracies for the proposed dataset are shown in
Figure~\ref{fig:loso-results}.

\begin{table*}[!t]
\centering
\small
\caption{Leave-one-subject-out performance on the proposed HamNoSys
dataset and ASL Fingerspelling Dataset A (\%). Values are reported as
mean $\pm$ sample standard deviation across 15 held-out subjects for
the proposed dataset and five held-out users for ASL Dataset A.}
\label{tab:loso-results}
\renewcommand{\arraystretch}{1.08}
\setlength{\tabcolsep}{2.5pt}

\begin{tabular*}{\linewidth}
{@{\extracolsep{\fill}}lccc@{}}
\toprule
Model & Accuracy@1 & Accuracy@3 & Accuracy@5 \\
\midrule
\multicolumn{4}{@{}l}{\textit{Proposed HamNoSys dataset (15 subjects)}} \\
ResNet-18
& $45.38\pm7.48$
& $67.72\pm8.80$
& $75.74\pm8.35$ \\

ViT-B/16
& $45.22\pm6.92$
& $68.47\pm7.75$
& $76.49\pm7.18$ \\

GCN
& $43.49\pm7.75$
& $69.23\pm9.15$
& $78.58\pm8.25$ \\

XGBoost
& $39.66\pm6.87$
& $64.29\pm8.75$
& $74.29\pm8.22$ \\
\addlinespace
\multicolumn{4}{@{}l}{\textit{ASL Fingerspelling Dataset A (5 users)}} \\
ResNet-18
& $87.40\pm4.50$
& $96.32\pm1.21$
& $97.95\pm0.68$ \\

ViT-B/16
& $82.20\pm5.36$
& $95.23\pm1.56$
& $97.44\pm0.73$ \\

GCN
& $84.22\pm4.03$
& $95.86\pm1.31$
& $97.75\pm0.81$ \\

XGBoost
& $86.14\pm3.81$
& $96.00\pm0.96$
& $97.76\pm0.49$ \\
\bottomrule
\end{tabular*}

\vspace{0.8em}

\begin{tabular*}{\linewidth}
{@{\extracolsep{\fill}}lcccccc@{}}
\toprule
& \multicolumn{3}{c}{Weighted average}
& \multicolumn{3}{c}{Macro average} \\
\cmidrule(lr){2-4}
\cmidrule(l){5-7}
Model
& P & R & $F_1$
& P & R & $F_1$ \\
\midrule
\multicolumn{7}{@{}l}{\textit{Proposed HamNoSys dataset (15 subjects)}} \\
ResNet-18
& $46.80\pm7.77$
& $45.38\pm7.48$
& $43.92\pm7.56$
& $46.76\pm7.81$
& $45.39\pm7.47$
& $43.90\pm7.57$ \\

ViT-B/16
& $47.64\pm7.29$
& $45.22\pm6.92$
& $43.63\pm7.07$
& $47.56\pm7.31$
& $45.22\pm6.89$
& $43.59\pm7.05$ \\

GCN
& $44.15\pm8.00$
& $43.49\pm7.75$
& $42.14\pm7.61$
& $44.10\pm8.03$
& $43.49\pm7.73$
& $42.11\pm7.62$ \\

XGBoost
& $40.44\pm6.97$
& $39.66\pm6.87$
& $38.49\pm6.72$
& $40.39\pm7.00$
& $39.67\pm6.86$
& $38.47\pm6.73$ \\
\addlinespace
\multicolumn{7}{@{}l}{\textit{ASL Fingerspelling Dataset A (5 users)}} \\
ResNet-18
& $88.75\pm3.95$
& $87.40\pm4.50$
& $87.02\pm4.52$
& $88.74\pm3.82$
& $87.43\pm4.68$
& $87.03\pm4.55$ \\

ViT-B/16
& $84.48\pm4.70$
& $82.20\pm5.36$
& $81.35\pm5.60$
& $84.45\pm4.70$
& $82.08\pm5.60$
& $81.24\pm5.74$ \\

GCN
& $86.16\pm3.36$
& $84.22\pm4.03$
& $83.97\pm3.94$
& $83.00\pm6.75$
& $81.56\pm5.36$
& $80.85\pm6.31$ \\

XGBoost
& $88.08\pm2.88$
& $86.14\pm3.81$
& $85.71\pm3.70$
& $84.64\pm6.05$
& $83.95\pm5.22$
& $82.72\pm6.23$ \\
\bottomrule
\end{tabular*}

\vspace{0.25em}
P = precision; R = recall.
\end{table*}

\section{Conclusion}
\label{sec:conclusion}

A balanced handshape dataset grounded in the official HamNoSys 4
Handshapes Chart was presented, comprising 144,000 RGB images from 15
participants across 160 illustrated classes. RGB appearance baselines
were provided by ResNet-18 and ViT-B/16, while hand-landmark baselines
were provided by a GCN and XGBoost. Complementary references for
seen-participant and participant-disjoint recognition were provided by
the subject-dependent and LOSO protocols.

A substantial effect of evaluation protocol on recognition performance
was observed, and the 160-class inventory remained challenging under
participant-disjoint testing. Matched-model evaluations were also
conducted on synthetic LSWH100 and real ASL Dataset A. The dataset and
baselines are intended to support phonology-grounded research and the
development of transcription, recognition, and translation tools across
sign languages, including under-resourced settings where dictionaries may
be available but labelled datasets remain scarce.

\paragraph{Limitations.}
Few limitations should be noted. Data were collected from 15
university students in a controlled indoor environment using a single
RGB camera, and broader demographic, environmental, and sensor
variability was therefore not represented. The class inventory was
restricted to the 160 static, single-hand forms illustrated in the
non-exhaustive HamNoSys chart; dynamic transitions, two-handed
configurations, orientation, location, movement, and non-manual
components were not included. Reference correspondence was verified by an
operator with sign-language experience, but further annotation validation by
HamNoSys specialists may strengthen the resource. Finally, only four
baseline model families were evaluated. More diverse participants,
less-controlled acquisition settings, and advanced fine-grained models
should be investigated in future work.

\backmatter

\section*{Statements and Declarations}

\bmhead{Ethics approval}

The study was conducted collaboratively by the Variable Energy Cyclotron
Centre and Ramakrishna Mission Vivekananda Educational and Research
Institute under the ethical and administrative procedures mutually
established by the participating organisations. All procedures involving
human participants were conducted in accordance with the applicable
institutional requirements.

\bmhead{Consent to participate}

Written informed consent was obtained from all participants before data
collection. Participation was voluntary, and the study procedure and
intended research use of the recordings were explained before the
recording sessions.

\bmhead{Consent for publication}

Written consent was obtained for the publication of participant images.
All participant faces shown in this article were anonymised before
publication.

\bmhead{Data availability}

The dataset will be made available after publication upon reasonable
request to the corresponding author at
\href{mailto:u.sarkar@vecc.gov.in}{u.sarkar@vecc.gov.in}. Access will be
subject to the conditions established by the collaborating institutions.
\bibliography{references}

\end{document}